\documentclass{article} 
\usepackage{src/collas2025_conference,times}
\usepackage{easyReview}
\usepackage{xcolor}
\usepackage{enumitem}

\usepackage{hyperref}
\hypersetup{
    colorlinks=true,
    linkcolor=red,
    filecolor=magenta,
    urlcolor=blue,
    citecolor=purple,
    pdftitle={Overleaf Example},
    pdfpagemode=FullScreen,
    }

\title{On the Hardness of Unsupervised Domain Adaptation: Optimal Learners and Information-Theoretic Perspective}
\preprintcopy
\author{
Zhiyi Dong\thanks{Equal contribution\\This paper has been accepted at the Conference on Lifelong Learning Agents 2025.} \\
University of Ottawa \\
\texttt{zdong014@uottawa.ca}
\And
Zixuan Liu\footnotemark[1] \\
University of Ottawa \\
\texttt{zliu282@uottawa.ca}
\And
Yongyi Mao \\
University of Ottawa \\
\texttt{ymao@uottawa.ca}
}

\usepackage{times}

\usepackage{mathrsfs}
\usepackage{bbm}
\usepackage{centernot}

\usepackage{amsmath, amssymb, amsthm}
\DeclareMathAlphabet{\mathrmt}{OT1}{ptm}{m}{n}

\usepackage{algorithmic}

\newcommand{\xs}{\text{\bf x}_\text{s}}
\newcommand{\xt}{\textbf{\text{\bf x}}_\text{t}}
\usepackage{wrapfig}
\usepackage{tikz}  
\usepackage{multirow}
\usetikzlibrary{arrows.meta, automata,
                positioning,
                quotes}

\newtheorem{thm}{Theorem}
\newtheorem{lem}{Lemma}
\newtheorem{rem}{Remark}

\newtheorem{coro}{Corollary}

\begin{document}

\maketitle

\begin{abstract}
This paper studies the hardness of unsupervised domain adaptation (UDA) under covariate shift. We model the uncertainty that the learner faces by a distribution $\pi$ in the ground-truth triples $(p, q, f)$\textemdash{}which we call a UDA class\textemdash{}where $(p, q)$ is the source-target distribution pair and $f$ is the classifier.  We define the performance of a learner as the overall target domain risk, averaged over the randomness of the ground-truth triple. This formulation couples the source distribution, the target distribution and the classifier in the ground truth, and deviates from the classical worst-case analyses, which pessimistically emphasize the impact of hard but rare UDA instances. In this formulation, we precisely characterize the optimal learner.  The performance of the optimal learner then allows us to define the learning difficulty for the UDA class and for the observed sample. To quantify this difficulty, we introduce an information-theoretic quantity\textemdash{}Posterior Target Label Uncertainty (PTLU)\textemdash{}along with its empirical estimate (EPTLU) from the sample , which capture the uncertainty in the prediction for the target domain. Briefly, PTLU is the entropy of the predicted label in the target domain under the posterior distribution of ground-truth classifier given the observed source and target samples.
By proving that such a quantity serves to lower-bound the risk of any learner, we suggest that these quantities can be used as proxies for evaluating the hardness of UDA learning. We provide several examples to demonstrate the advantage of PTLU, relative to the existing measures, in evaluating the difficulty of UDA learning. 

\end{abstract}

\section{Introduction}
Unsupervised domain adaptation (UDA) is a learning methodology that exploits a labeled sample in the ``source domain'' and an unlabeled sample in the ``target domain'' to construct a classifier for the target domain. Such an approach is of great practical significance when labeled examples are difficult or expensive to acquire.  In the deep learning era, numerous UDA learning algorithms (see, e.g., \cite{DANN, liu2016coupled, shen2018wasserstein, song2020bridging, de2023value}) have been developed and demonstrated to be effective on various benchmark datasets. However, these algorithms are mostly designed heuristically, with the objective of aligning the target distribution with the source distribution (possibly in a latent space) using a well-defined loss function.

Theoretical analysis of UDA has a rich body of literature. One class of approaches focuses on quantities that measure the difference or discrepancy between the source domain distribution $p$ and the target domain distribution $q$\textemdash{}without considering the hypothesis class used for learning\textemdash{}and derive learning bounds based on these quantities. Examples of such quantities include KL divergence \citep{nguyen2021kl, wang2022information}, Kernel Mean Matching \citep{huang2006correcting}, Rényi divergence \citep{cortes2010learning, mansour2012multiple, hoffman2018algorithms}, among others. Leaving the hypothesis classes out of consideration, such approaches are clearly inadequate in capturing the structure of UDA problems. The pioneering work of \cite{HdeltaH} (an extended version of \cite{ben2006analysis} and \cite{blitzer2007learning}) is the first to leap from this paradigm, where the hypothesis class is explicitly brought into analysis and a discrepancy measure between $p$ and $q$ with respect to a hypothesis class ${\cal H}$ is introduced, under the name ${\cal H}\Delta {\cal H}$ divergence. This perspective has inspired various notions of other discrepancy measures that also consider the hypothesis class. Examples of such measures include \cite{YDiscp}, \cite{mansour2009domain}, \cite{kuroki2019unsupervised}, and \cite{shui2022novel}. Particularly noteworthy is the notion of transfer exponent \citep{TransExp}, which overcomes certain limitations of the previous measures (e.g., symmetry between $p$ and $q$).  These measures usually lead to statistical learning bounds and often inspire practicable learning algorithms (e.g. in \cite{zhang2019bridging}, \cite{acuna2021f}, and \cite{wang2025f}). 

Despite these theoretical advances, the discrepancy measures proposed to date remain inadequate to fully capture the inherent structure of UDA learning problems. For example, the learning bounds in \cite{ben2006analysis}, \cite{blitzer2007learning}, \cite{HdeltaH}, \cite{TransExp}, and \cite{wang2025f} are overly pessimistic, as they express the performance of the worst member in the hypothesis class. On the other hand, most of the existing theoretical studies set out to study a fixed pair $(p, q)$ of source and target distributions and a fixed hypothesis class ${\cal H}$. Such a formulation does not capture the fact that the ground-truth source and target distributions $(p, q)$ are unknown to the learner, and each possible $(p, q)$ pair may be coupled with a different family of ground-truth classifiers.\textemdash{}Indeed, the work of \cite{ben2012hardness} already hints that this coupling may severely impact the difficulty of learning.

In this paper, we study the hardness of UDA (under covariate shift) using a different formulation. Specifically, we model the uncertainty that the learner faces by a distribution $\pi$ of the ground-truth triples $(p, q, f)$\textemdash{}which we call a UDA class\textemdash{}where $(p, q)$ is the source-target distribution pair and $f$ is the classifier. When Nature picks $(p, q, f)$ from $\pi$ and reveals the labeled source sample and unlabeled target sample, the learner (possibly stochastic) must pick a classifier and use it to classify the input drawn from $q$. Instead of using the worst-case risk on the target domain, we consider as the performance metric the overall target domain risk, averaged over the random choices of Nature. The advantages of this formulation are two-fold. First, it naturally couples the ground-truth triple: the source distribution, the target distribution and the classifier. Second, this formulation deviates from the worst-case analyses, de-emphasizing the impact of hard but rare UDA instances on the performance of the learner. 

In this setting, we show that the overall risk of the learner decomposes into (the weighted sum of) its ``sample-wise" risks associated with each UDA instance $(p, q, f)$. This decomposition suggests that the optimal learner measured under overall risk must also behave optimally with respect to the sample-wise risk.  We then characterize the exact form of the optimal learner and use the optimal overall risk to measure the learning difficulty associated with the UDA class $\pi$ and use the optimal sample-wise risk to measure the difficulty on learning with each observed UDA sample.

We then introduce the notion of Posterior Target Label Uncertainty (PTLU), an information-theoretic notion, as a proxy for learning difficulties. In essence, PTLU measures the average amount of uncertainty in predicting the label for inputs drawn from the target domain based on the posterior distribution classifier conditioned on the observed sample. Invoking Fano's inequality \citep{fano1952class, cover1999elements}, we show that PTLU serves as a critical quantity in lower-bounding the sample-wise risk\textemdash{}and hence the overall risk\textemdash{}of any learner. We also prove a high-probability lower-bound on target-domain risk for any classifier consistent with the source samples, exploiting the notion of PTLU. We present several examples to demonstrate the advantage of PTLU upon other discrepancy measures in evaluating the hardness of UDA learning.

Since computing the optimal sample-wise risk and PTLU requires the precise knowledge of the target distribution $q$, we introduce an empirical version of PTLU (EPTLU), which can be directly measured using the target sample without access to $q$, and prove a high-probability lower bound on the sample-wise risk based on EPTLU. Furthermore, by proving another high-probability lower bound based on EPTLU, we show that one can estimate the learning difficulty in the UDA class $\pi$ from an infinite-sample UDA observation based on EPTLU computed from a finite sample. 

The main contributions of this work are summarized as follows.
\begin{itemize}
\item We present a novel probabilistic formulation of UDA problems, de-emphasizing the influence of rare instances. 

\item In this formulation, we derive the optimal learner and use the optimal risk to measure the hardness of UDA learning.

\item We prove a risk lower bound in terms of 
an information-theoretic quantity, which we refer to as the Posterior Target Label Uncertainty, or PTLU.

\item We show via concrete examples that PTLU can be used as a proxy of hardness in UDA learning, and demonstrate its superiority to various existing metrics. 

\end{itemize}

The proofs and computational procedures of all results are included in Appendix, and Table \ref{tab:notations} summarizes the notations used throughout this work.

\section{Problem Formulation}
With rare exceptions, we will adopt the notational convention where we use a calligraphic capitalized letter, say, $\mathcal{X}$,  to denote a set, and its corresponding lower-cased letter, in this case, $x$, to denote an element in the set.
For any measurable set $\mathcal{A}$, we denote by $\Delta(\mathcal{A})$ the set of all probability distributions on $\mathcal{A}$. For any distribution $\mu \in \Delta(\mathcal{A})$, we will use $\Omega(\mu)$ to denote the support of $\mu$, and $\mu^n$ to denote the distribution in $\Delta(\mathcal{A}^n)$ defined by $\mu^n(a_1, a_2, \ldots, a_n)=\prod_{i=1}^n \mu(a_i)$. If $A$ is a measurable subset of $\mathcal {A}$, we will, with a slight abuse of notation, write $\mu(A)$ in place of the probability of the event $A$ under $\mu$, namely, the value $\int_{a\in A} \mu(a) da$. That is, we will make no distinction between a probability measure and a probability distribution in our notations.  We will also make use of random variable notations, which will be denoted consistently by capitalized letters, e.g., $X$, $Y$. If $\mu$ is a distribution in $\Delta(\mathcal {X}\times \mathcal {Y})$, it corresponds to a unique pair $(X, Y)$ of random variables. Then we may write $\mu$ alternatively as $\mu_{XY}$. In addition, the marginal distribution of $X$ and the conditional distribution of $Y$ conditioned on $X$, both induced from $\mu_{XY}$, are well defined, which will be denoted by $\mu_{X}$  and $\mu_{Y\vert X}$ respectively.

The learning problem we consider will have input space $\mathcal{X}$ and output space $\mathcal{Y}$. We will primarily consider the setting where $\mathcal{Y}$ is finite\textemdash{}corresponding to a classification setting\textemdash{}and $\mathcal{X}$ is a measurable subset of ${\mathbb R}^d$. We will use $\mathcal{F}$ to denote the space of all measurable functions mapping $\mathcal{X}$ to $\mathcal{Y}$, and each such function is referred to as a {\em hard classifier}.  This choice of terminology is to contrast the notion of a {\em soft classifier}, which is a measurable function mapping ${\cal X}$ to the space $\Delta({\cal Y})$ of all distributions on ${\cal Y}$. We denote the space of all such soft classifiers by ${\cal F}^{\rm s}$. It is clear that each hard classifier in ${\cal F}$ can be identified with a soft classifier in ${\cal F}^{\rm s}$. For a soft classifier $t\in {\cal F}^{\rm s}$, we will use $t(y\vert x)$ to denote the probability of label $y$  under the distribution $t(x)$. Concerning soft classifiers and hard classifiers, we would like to highlight the following two simple facts.  First, any soft classifier $t\in {\cal F}^{\rm s}$ can be ``hardened'' to a hard classifier $t^{\rm H}\in {\cal F}$ via
\begin{equation}
    t^{\rm H}(x):=\arg\max_{y'\in {\cal Y}} \; t(y'\vert x).
\end{equation}
Second, any distribution $\mu$ on the space ${\cal F}$ of hard classifiers (i.e., $\mu \in \Delta({\cal F})$) induces a soft classifier $\mu^{\rm A} \in {\cal F}^s$ via
\begin{equation}
    \mu^{\rm A}(y\vert x):= {\rm Pr}_{g\sim \mu}\big(g(x)=y\big)
\end{equation}
and we refer to $\mu^{\rm A}$ as the {\em aggregated (soft) classifier} from $\mu$. From here on, we may refer to hard classifiers simply as ``classifiers'' for succinctness.

With these notations, we now formulate the unsupervised domain adaptation (UDA) problem.

{\bf Model of Nature.} Central to this formulation is a distribution $\pi$ over $\Delta(\mathcal{X})\times\Delta(\mathcal{X})\times\mathcal{F}$. We call $\pi$ a {\em UDA class}. From this distribution $\pi$, Nature draws a triple $(p,q,f)$, where $p$ is an input distribution in the source domain, $q$ is an input distribution in the target domain, and $f$ is a classifier. We assume that the source and target domains share the same classifier $f$, namely, the setting of covariate shift \citep{CovariateShift}. We call the triple $(p, q, f)$ a {\em UDA instance} in which the three elements specify the ground-truth.

After picking the UDA instance $(p, q, f)$, Nature then draws a vector ${\bf x}_{\rm s}:=\big(x_1,\cdots,x_m\big)$ from $p^m$, and creates the labels of its element as vector $\big(f(x_1),\cdots,f(x_m)\big)$, which we denote by $f({\bf x}_{\rm s})$. The pair $\big({\bf x}_{\rm s}, f({\bf x}_{\rm s})\big)$ then forms the labeled source sample of size $m$. Nature also draws a vector ${\bf x}_{\rm t}:=\big(x_1',\cdots,x_n'\big)$ from $q^n$, serving as the unlabeled target sample of size $n$. The source-target sample $\big({\bf x}_{\rm s},  {\bf x}_{\rm t}, f({\bf x}_{\rm s})\big)$ is then revealed to the learner.  For convenience, we will denote $\mathcal{S}_{m, n}:=\mathcal {X}^m \times \mathcal{X}^n \times \mathcal {Y}^m$. Each element of $\mathcal {S}_{m, n}$, such as $\big({\bf x}_{\rm s},  {\bf x}_{\rm t}, f({\bf x}_{\rm s})\big)$, is referred to as an $(m, n)$-sample. When an $(m, n)$-sample $s_{m, n}: =\big({\bf x}_{\rm s},  {\bf x}_{\rm t}, {\bf y}_{\rm s}\big)$ has its ${\bf x}_{\rm s}$ drawn from $p^m$, ${\bf x}_{\rm t}$ drawn from $q^n$ and ${\bf y}_{\rm s}=f({\bf x}_{\rm s})$, we write $s_{m, n}\sim (p, q, f)$, and say that {\em $s_{m,n}$ is an $(m, n)$-realization of $(p, q, f)$.}

{\bf Model of Learner.} The learner has no prior knowledge of the UDA instance $(p, q, f)$. Its only prior knowledge is the distribution $\pi$. The behaviour of the learner is completely characterized by its learning algorithm, namely, a function $\mathscr{A}_{m, n}: \mathcal {S}_{m, n}\rightarrow \Delta (\mathcal {F})$ mapping the space of all $(m, n)$-samples to the space of all distributions on $\mathcal {F}$. Specifically, upon observing an $(m, n)$-sample $s_{m, n} =\big({\bf x}_{\rm s},  {\bf x}_{\rm t}, {\bf y}_{\rm s}\big)$ that is a realization of $(p, q, f)$ (but $(p, q, f)$ is unknown to the learner), the learner draws a hard classifier $g$ from the conditional distribution $\mathscr{A}_{m, n}(\cdot\vert s_{m, n})$, and uses $g$ to label any input drawn from $q$.

In the limit when $m$ and $n$ go to infinity, observing the sample $s_{m,n}$ is equivalent to observing the source distribution $p$, the target distribution $q$ and the restriction of $f$ on the support $\Omega(p)$ of the source distribution $p$. For simplicity, we denote by $f_p$ the restriction of $f$ on $\Omega(p)$. Then in the limit of  infinite sample sizes $m$ and $n$, we write the algorithm $\mathscr{A}_{m, n}$ as $\mathscr{A}_{\infty}$, its input as $(p, q, f_p)$, and its output as the distribution $\mathscr{A}_{\infty}(\cdot\vert p, q, f_p)$ on  $\Delta({\cal F})$. We refer to the triple $(p, q, f_p)$ as an {\em infinite-sample UDA observation}.

We note that this formulation of the learner's knowledge differs from that in PAC-learning \citep{PACLearning} for classical supervised learning in two ways. First,  the analogue to the notion of concept in PAC-learning is extended to the notion of UDA instance.  This treatment is in line with that of \cite{ben2012hardness} and \cite{ben2014domain}. Second, instead of considering a set of UDA instances as in \cite{ben2012hardness} and \cite{ben2014domain}, the knowledge of the learner is modelled as a distribution over the space of UDA instances. The idea of modelling learner's uncertainty, knowledge and bias about the ground-truth using a distribution is not new\textemdash{}it has been exploited in various contexts, for example in PAC-Bayes \citep{mcallester1998some, shawe1997pac} and meta-learning \citep{baxter2000model}.

{\bf Performance Metrics.} We now define the performance metric of the learner. We will overload the notation $R(\cdot)$ to denote various kinds of risk, which should be distinguishable from the argument inside the parentheses.

For any UDA instance $(p,q,f)$, the {\em  target domain risk of a classifier $g\in \mathcal {F}$} is defined as
\begin{equation}
  \label{eq:risk_g_qf}
    R(g\vert q, f):=\mathbb{E}_{x\sim q}
    \mathbbm{1}\{
    g(x)\neq f(x)\}.
\end{equation}
Note that this risk is independent of the source domain input distribution $p$. The {\em expected target domain risk of the learner $\mathscr{A}_{m, n}$} for the UDA instance $(p, q, f)$ is then
\begin{equation}
  \label{eq:risk_A_pqf}
R(\mathscr{A}_{m, n}\vert p, q, f):=
\mathbb{E}_{s_{m, n}\sim (p, q, f)} \mathbb{E}_{g\sim \mathscr{A}_{m, n}(\cdot\vert s_{m, n})}
R(g\vert q, f)
\end{equation}
Notably, in asymptotic regime when $m$ and $n$ go to infinity, this risk becomes
\begin{equation}
R(\mathscr{A}_{\infty}\vert p, q, f) :=\mathbb{E}_{g\sim \mathscr{A}_{\infty}(\cdot\vert p, q, f_p)} R(g\vert q, f)
\end{equation}
Finally, the {\em overall target domain risk} of the learner $\mathscr{A}_{m, n}$ on the UDA class $\pi$ is defined as
\begin{equation}
\label{eq:risk_pi}
R(\mathscr{A}_{m, n}): = \mathbb{E}_{(p, q, f)\sim \pi} R(\mathscr{A}_{m, n}\vert p, q, f)
\end{equation}
In this paper, the objective of the learner is to achieve a small overall target domain risk defined in (\ref{eq:risk_pi}). The learner that minimizes this risk is referred to as the {\em optimal learner} $\mathscr{A}^*_{m,n}$. The risk $R(\mathscr{A}^*_{m, n})$ of the optimal learner is denoted by $R^*_{m, n}$. We then use $R^*_{m, n}$ to measure the difficulty of learning UDA class $\pi$ with an $(m, n)$-sample, and use $R^*_{\infty}$ to measure the difficulty of learning with infinite-sample observations.

\section{Optimal Learner, Optimal Risks, and Risk Lower Bounds}
\subsection{Posterior Distribution of Classifier}
For any ${\bf x}=(x_1, x_2, \ldots,  x_m)\in \mathcal {X}^m$ and any ${\bf y}=(y_1, y_2, \ldots, y_m)\in \mathcal {Y}^m$, let 
\begin{equation}
    \mathcal{F}[{\bf x}, {\bf y}]:=\{g\in\mathcal{F}:g({\bf x})={\bf y}\},
\end{equation}
that is, $\mathcal {F}[{\bf x}, {\bf y}]$ is the set of all classifiers that label ${\bf x}$ as ${\bf y}$. The equivalent notion to $\mathcal{F}[{\bf x}, {\bf y}]$ in the limit of infinite $m$ is ${\cal F}[f_p]$,  defined by
\begin{equation}
    \mathcal{F}[f_p]:=\{g\in\mathcal{F}:g(x)=f_p(x), \forall x\in {\rm dom}f_p\},
\end{equation}
where ${\rm dom}f_p$ denotes the domain of function $f_p$ and is in fact $\Omega(p)$.  Then $\mathcal{F}[f_p]$ is the set of all classifiers that are consistent with $f_p$ on the support of $p$.

We now treat $\pi$ as the distribution of random variable triplet $(P, Q, F)$ and rewrite it as $\pi_{PQF}$. When a random $(m, n)$-sample $S$ realizes $(P, Q, F)$, the distribution $\pi_{PQF}$ induces the joint distribution  $\pi_{PQFS}$ of $(P, Q, F, S)$ via
\begin{equation}
\pi_{PQFS}{(p, q, f, s_{m, n}}) = 
\pi_{PQF}(p, q, f) p^m({\bf x}_{\rm s}) q^n({\bf x}_{\rm t})
\mathbbm{1}\{f\in \mathcal {F}[{\bf x}_{\rm s}, {\bf y}_{\rm s}]\}.
\end{equation}
Under $\pi_{PQFS}$, the conditional distribution $\pi_{PQF\vert S}$ of $(P, Q, F)$ conditioned on $S$ is well defined by
\begin{equation}
\pi_{PQF\vert S}(p, q, f\vert s_{m, n}) = \frac{\pi_{PQFS}{(p, q, f, s_{m, n}})}{\pi_S(s_{m, n})}
=\frac{\pi_{PQFS}{(p, q, f, s_{m, n}})}{\int\pi_{PQFS}{(p', q', f', s_{m, n}})dp'dq'df'}.
\end{equation}
Then the conditional distribution $\pi_{F\vert S}$ of $F$ conditioned on $S$ is defined by
\begin{equation}
    \pi_{F\vert S}(f\vert s_{m,n})=\int\pi_{PQF\vert S}(p,q,f\vert s_{m,n})dpdq := {\rho}(f\vert s_{m,n}).
\end{equation}
This posterior distribution over the space of classifiers given a observed $(m, n)$-sample appears to play the central in the study UDA problem. For this reason, we also dedicate to it a special notation $\rho(\cdot\vert \cdot)$.

We now define the same notion of posterior distribution for  infinite $m$ and $n$. To that end, first note that the conditional distribution of $F$ conditioned on $(P,Q)$ is
\begin{equation}
    \pi_{F\vert P,Q}(f\vert p,q)
    =\dfrac
    {\pi_{P,Q,F}(p,q,f)}
    {\pi_{P,Q}(p,q)}
    =\dfrac
    {\pi_{P,Q,F}(p,q,f)}
    {\int\pi_{P,Q,F}(p,q,f')df'}.
\end{equation}
Let $\hat{f}$ be a function mapping $\Omega(p)$ to ${\cal Y}$.  Under distribution $\pi_{PQF}$, the conditional distribution of $F$ conditioned on 
$P=p, Q=q$ and $F_p=\hat{f}$, which we will denote by $\rho_\infty(\cdot\vert p, q, \hat{f})$, is then
\begin{equation}
\rho_\infty(f\vert p, q, \hat{f}):= \dfrac
    {\pi_{F\vert P,Q}(f\vert p,q)\mathbbm{1}\{f_p=\hat{f}\}}
    {\int\pi_{F\vert P,Q}(f'\vert p,q)\mathbbm{1}\{f'_p=\hat{f}\}df'}.
\end{equation}
Clearly, the function  $\rho_\infty(\cdot\vert p, q, \hat{f})$ is well defined for every $(p, q, \hat{f})$. Then the function  $\rho_\infty(\cdot\vert p, q, f_p)$ is well defined. It is easy to verify (as shown by Eq. (\ref{eq:rho finite vs infinite}) and proved in Appendix \ref{proof:thm:finite to infinite})  that 
when $s_{m, n}\sim (p, q, f)$, $\rho(\cdot\vert s_{m,n})$ converges to $\rho_\infty(\cdot\vert p, q, f_p)$ as $m$ and $n$ go to infinity. That is, $\rho_\infty(\cdot\vert p, q, f_p)$ is essentially the posterior distribution over the space of classifiers conditioned on an infinite-sample UDA observation $(p, q, f_p)$. For this reason, we will write $\rho(\cdot\vert p, q, f_p)$ in place of $\rho_{\infty}(\cdot\vert p, q, f_p)$. That is, we will overload the notation $\rho(\cdot\vert \cdot)$ to refer to both $\rho(\cdot\vert s_{m,n})$ and $\rho(\cdot\vert p, q, f_p)$. 

\subsection{Optimal Learners}
Given the posterior distribution $\rho(\cdot\vert \cdot)$, its aggregated soft classifier $\rho^{\rm A}(\cdot\vert \cdot)$ is well defined. Specifically,  we will use $\rho^{\rm A}(y\vert x, s_{m, n})$ to  denote the probability that the aggregated classifier from $\rho(\cdot\vert s_{m, n})$ assigns label $y$ to $x$. Likewise,  $\rho^{\rm A}(y\vert x, p, q, f_p)$ denotes the probability that the aggregated classifier from $\rho(\cdot\vert p, q, f_p)$ assigns label $y$ to input $x$. The hardened classifiers $\left(\rho^{\rm A}\right)^{\rm H}(\cdot\vert s_{m, n})$ and $\left(\rho^{\rm A}\right)^{\rm H}(\cdot\vert p, q, f_p)$  of $\rho^{\rm A}(\cdot\vert \cdot, s_{m, n})$ and $\rho^{\rm A}(\cdot\vert \cdot, p, q, f_p)$ are also well defined. 

For any $(m, n)$-sample $s_{m, n}$ realizing an UDA instance $(p, q, f)$, we now define the risk of learner $\mathscr{A}_{m, n}$ when learning on sample $s_{m, n}$.
\begin{equation}
    e(\mathscr{A}_{m, n}; s_{m,n}, q):=\mathbb{E}_{x\sim q} \left\{\underset{\substack{f\sim \rho(\cdot\vert s_{m, n}) \\ g\sim\mathscr{A}_{m, n}(\cdot\vert s_{m, n})}}{\Pr}\big(f(x)\neq g(x)\big)\right\},
\end{equation}
which is referred to as the {\em sample-wise risk} of the learner $\mathscr{A}_{m, n}$. 
For an infinite-sample UDA observation $(p, q, f_p)$, the sample-wise risk is likewise defined as
\begin{equation}
    e(\mathscr{A}_{\infty}; p, q, f_p):=\mathbb{E}_{x\sim q}    \left\{\underset{\substack{f\sim \rho(\cdot\vert p, q, f_p) \\ g\sim\mathscr{A}_{\infty}(\cdot\vert p, q, f_p) }}{\Pr}\big(f(x)\neq g(x)\big)\right\}.
\end{equation}
Then, the following results are obtained by the decomposition of the overall target domain risk for any learner, which is derived in Appendix \ref{apx:decp}.
\begin{lem} 
\label{lem:key}
For any learner $\mathscr{A}_{m, n}$ and $\mathscr{A}_{\infty}$,
\begin{equation}
    R(\mathscr{A}_{m, n}) = {\mathbb E}_{(p, q, f) \sim \pi}
    {\mathbb E}_{s_{m, n}\sim (p, q, f)}
    e(\mathscr{A}_{m, n}; s_{m,n}, q), 
\end{equation}
and
\begin{equation}
    R(\mathscr{A}_{\infty})={\mathbb E}_{(p, q, f) \sim \pi} 
    e(\mathscr{A}_{\infty}; p, q, f_p).
\end{equation}
\end{lem}
This simple lemma reveals an important perspective in understanding the overall risk of a learner, namely, that it decomposes into sample-level risks each measuring the discrepancy between the learner's output distribution over ${\cal F}$ for an observed sample and the posterior distribution over ${\cal F}$ conditioned on the sample. Thus, if a learner is optimal for every sample, it is optimal overall for the UDA class $\pi$. To that end, we define the optimal sample-wise risk as
\begin{equation}
e^*(s_{m, n}, q):= \min_{\mathscr{A}_{m, n}} e(\mathscr{A}_{m, n}; s_{m, n}, q)  ~~{\rm and }~ ~
e^*(p, q, f_p):= \min_{\mathscr{A}_{\infty}} e(\mathscr{A}_{\infty}; p, q, f_p).
\end{equation}

\begin{thm}
\label{thm:optimal_mn}
    The optimal learner $\mathscr{A}^*_{m,n}$ returns the hard classifier $\left(\rho^{\rm A}\right)^{\rm H}(\cdot\vert s_{m,n})$ with probability $1$ on every $(m, n)$-sample $s_{m,n}$ realizing some $(p, q, f)$, and
    \begin{equation}
        e^*(s_{m, n}, q) = 1-  
        \mathbb{E}_{x\sim q} 
        \max_{y} \rho^{\rm A}(y\vert x, s_{m, n}).
    \end{equation}
\end{thm}

 \begin{coro}
 \label{coro:optimal_infty}       
The optimal learner $\mathscr{A}^*_{\infty}$ returns the hard classifier $\left(\rho^{\rm A}\right)^{\rm H}(\cdot\vert p, q, f_p)$
    with probability $1$ on every infinite-sample UDA observation $(p,q,f_p)$, and
\begin{equation}
        e^*(p, q, f_p) = 1- 
        \mathbb{E}_{x\sim q} 
        \max_{y} \rho^{\rm A}(y\vert x, p, q, f_p).
    \end{equation}
\end{coro}    

Theorem \ref{thm:optimal_mn} and Corollary \ref{coro:optimal_infty} (proved in Appendix \ref{apx:optimal}) allow us to immediately obtain $R^*_{m,n}$ and $R^*_{\infty}$ for a given UDA class (using Lemma \ref{lem:key}). In addition, they suggest that $e^*(s_{m, n}, q)$ and $e^*(p, q, f_p)$ can be used to measure the difficulty of learning from sample $s_{m, n}$ and $(p, q, f_p)$ respectively.

\subsection{Risk Lower Bounds from Posterior Target Label Uncertainty}
Let $\mathscr{H}(\cdot)$ denote the entropy functional. The Posterior Target Label Uncertainty (PTLU) on an $(m, n)$-sample $s_{m,n}$ realizing an UDA instance $(p, q, f)$ is defined as
\begin{equation}
U(s_{m,n}, q) := \mathbb{E}_{x\sim q} \mathscr{H}\left(\rho^{\rm A}(\cdot\vert x, s_{m,n})\right),
\end{equation}
and the PTLU on an infinite-sample UDA observation $(p, q, f_p)$ is defined as
\begin{equation}
U(p, q, f_p) := \mathbb{E}_{x\sim q} \mathscr{H}\left(\rho^{\rm A}(\cdot\vert x, p,q,f_p)\right).
\end{equation}

We note that the quantity $U(\cdot)$ also depends on $\pi$, since $\rho(\cdot\vert \cdot)$ depends on $\pi$. The quantity measures the average amount of uncertainty in the prediction by the aggregated soft classifier $\rho^{\rm A}(\cdot\vert \cdot)$. 

\begin{thm}
\label{thm:LB_mn}
Assume that $ \mid\mathcal{Y}\mid =k$. For any learner $\mathscr{A}_{m,n}$ and any sample $s_{m, n}$ realizing an UDA instance $(p, q, f)$:
    \begin{equation}
    \begin{aligned}
    \label{eq:LB_mn}
       k>2:  e(\mathscr{A}_{m,n}; s_{m, n}, q)&\geq \frac{U(s_{m,n}, q)-1}{\log (k-1)},\\
        k=2:  e(\mathscr{A}_{m,n}; s_{m, n}, q)&\geq \frac{U^2(s_{m,n}, q)}{4}+e^*(s_{m,n}, q)^2.
    \end{aligned}
    \end{equation}
\end{thm}

\begin{coro}
\label{coro:LB_infty}
Assume that $\mid\mathcal{Y}\mid=k$. For any learner $\mathscr{A}_{\infty}$ and any infinite-sample UDA observation 
$(p, q, f_p)$:
    \begin{equation}
    \begin{aligned}
    \label{eq:LB_infty}
        k>2:   e(\mathscr{A}_{\infty}; p, q, f_p)&\geq \frac{U(p,q,f_p)-1}{\log (k-1)},\\
        k=2:  e(\mathscr{A}_{\infty}; p, q, f_p)&\geq \frac{U^2(p,q,f_p)}{4}+e^*(p,q,f_p)^2.
    \end{aligned}
    \end{equation}
\end{coro}

For any single classifier $g$, consistent with source sample, Theorem \ref{thm:R(g)} shows that PTLU also serves in a high-probability lower bound of its target domain risk.

\begin{thm}
\label{thm:R(g)}
    Let $s_{m,n}$ realize a UDA instance $(p,q,f)$. For any $g\in \mathcal{F}[{\bf x}_\text{s},{\bf y}_\text{s}]$, with probability at least $1-\delta$, 
\begin{equation}
\begin{aligned}
        k>2: \;\;& R(g\vert q,f)\geq \frac{U(s_{m,n},q)-1}{\log(k-1)}-\sqrt{\frac{\mathbb{V}_{f\sim \rho(\cdot\vert s_{m,n})}[R(g\vert  q,f)]}{\delta}},\\
        k=2: \;\;&R(g\vert q,f) \geq \frac{U^2(s_{m,n},q)}{4} +e^*(p,q,f_p)^2-\sqrt{\frac{\mathbb{V}_{f\sim \rho(\cdot\vert s_{m,n})}[R(g\vert  q,f)]}{\delta}},
\end{aligned}
\end{equation}
where $\mathbb{V}$ denotes the variance operator.
\end{thm}

\subsection{Examples and Comparison with Other Discrepancy}
We now use PTLU to evaluate learning difficulty in four case studies, each assuming binary classification (i.e., $\mathcal{Y}=\{0,1\}$) . We compare PTLU with several widely-used measures, including:  
\begin{enumerate}
    \item $f$-divergence \citep{csiszar1963information} between $p$ and $q$: $D_f(p\Vert q)=\mathbb{E}_{x\sim q}\left[f\left(p(x)/q(x)\right) \right]$ with a convex function $f$ satisfying $f(1)=0$; 
    \item  Wasserstein distance \citep{panaretos2019statistical} between $p$ and $q$: $W^d(p\Vert q)=\inf_{x_1\sim p, x_2\sim q} \mathbb{E}^{1/d}\Vert x_1-x_2\Vert^d$ with parameter $d\geq 1$;
    \item $\mathcal{H}\Delta\mathcal{H}$ divergence \citep{HdeltaH} between $p$ and $q$ for a hypothesis class $\mathcal{H}\subseteq \mathcal{F}$: $d_{\mathcal{H}\Delta\mathcal{H}}(p\Vert q)=\sup_{h,h'\in\mathcal{H}}\left|\Pr_{x\sim p} (h(x)\neq h'(x))-\Pr_{x\sim q} (h(x)\neq h'(x))\right|$;
    \item $\mathcal{Y}$-discrepancy \citep{YDiscp} for a UDA instance $(p,q,f)$ and a hypothesis class $\mathcal{H}\subseteq \mathcal{F}$: $d_{\mathcal{Y}}(p\Vert q)=\sup_{h\in\mathcal{H}}\left|\Pr_{x\sim p} (h(x)\neq f(x))-\Pr_{x\sim q} (h(x)\neq f(x))\right|$;
    \item Marginal transfer exponent $\gamma$ \citep{TransExp} for a UDA instance $(p,q,f)$ with respect to a hypothesis class $\mathcal{H}\subseteq \mathcal{F}$:  if there exists $C_\gamma$ such that for all  $h\in\mathcal{H}$, $C_\gamma \Pr_{p} (h(X)\neq f(X))\geq \Pr_{q} ^{\gamma}(h(X)\neq f(X))$ for some $\gamma>0$,  then $\gamma$ is called the marginal transfer exponent. 
\end{enumerate}

For each example, we proceed as follows: (1) we first specify the involved UDA class $\pi$;  (2) we then compute the optimal overall target domain risk $R^*_{\infty}$ for $\pi$ (namely, assuming access to an infinite-sample observation), where a smaller value indicates easier learning; (3) next, we select the hardest UDA instance from each involved $\pi$ (measured in terms of optimal sample-wise risk $e^*(p, q, f_p)$), and compute the PTLU $U(p, q, f_p)$ along with other comparison  measures on the infinite-sample observation $(p, q, f_p)$; and (4) finally, we evaluate whether each measure accurately reflects the transfer difficulty. We focus on the hardest UDA instances because the $\mathcal{H}\Delta\mathcal{H}$ divergence, $\mathcal{Y}$-discrepancy, and marginal transfer exponent all reflect the hardest/worst-case classifier in $\mathcal{H}$.  Detailed calculations and a summary table of all computed measures are provided in Appendix \ref{apx:comp_examples}.

\textbf{Example 1.} 
We consider two UDA classes, $\pi^1$ and $\pi^2$, defined below. As we will show, $\pi^1$ is easier to transfer than $\pi^2$, as evidenced by a lower overall target domain risk $R^*_{\infty}$ of the optimal learner.  However, some traditional divergence-based measures fail to capture this distinction as they ignore the interaction between distributions and labeling functions. 

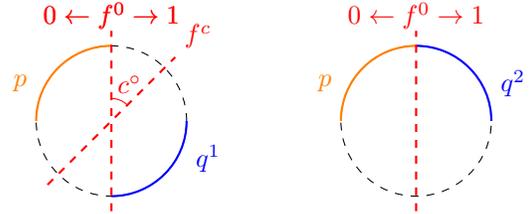
\begin{wrapfigure}{R}{0.5\textwidth}
   \begin{minipage}{0.24\textwidth}
        \centering         
        \begin{tikzpicture}
           \draw[black, dashed] (1,0) arc[start angle=0, end angle=360, radius=1cm];
            \draw[orange, thick] (-1,0) arc[start angle=180, end angle=90, radius=1cm];
    
            \draw[blue, thick] (0,-1) arc[start angle=270, end angle=360, radius=1cm];
    
            \draw[red, thick, dashed] (0,-1.2) -- (0,1.2);
            \node[above, red] at (0,1.1) { $0 \leftarrow f^0 \rightarrow 1$};
            \draw[red, thick, dashed] (-0.84852,-0.84852) -- (0.84852,0.84852);
            \node[above right, red] at (0.84852,0.84852) { $ f^c$};
            \draw[red] (0,0.3) arc[start angle=90, end angle=45, radius=0.3cm];
            \node[above, red] at (0.24,0.24) { $c^\circ$};
            \node[left, orange] at (-1,0.5) { $p$};
            \node[right, blue] at (1,-0.5) { $q^1$};
             \node[above, red] at (0,1.1) { $0 \leftarrow f^0 \rightarrow 1$};
        \end{tikzpicture}
   \end{minipage}
   \begin{minipage}{0.24\textwidth}
        \centering
        \begin{tikzpicture}
            \draw[black, dashed] (1,0) arc[start angle=0, end angle=360, radius=1cm];
            \draw[thick, orange] (-1,0) arc[start angle=180, end angle=90, radius=1cm];
            \draw[thick, blue] (0,1) arc[start angle=90, end angle=0, radius=1cm];
    
            \draw[red, thick, dashed] (0,-1.2) -- (0,1.2);
            
            \node[left, orange] at (-1,0.5) { $p$};
            \node[right, blue] at (1,0.5) { $q^2$};
            \node[above, red] at (0,1.1) { $0 \leftarrow f^0 \rightarrow 1$};
        \end{tikzpicture}
   \end{minipage}
   \caption{Distributions $p,q^1,q^2$ and classifiers $f^c$ of two UDA classes $\pi^1$ (left) and $\pi^2$ (right) in Example 1.}
   \label{Fig:eg2}
\end{wrapfigure}

Let $\Omega(\pi_{P,Q}^1):=\{(p,q^1)\}$ and  $\Omega(\pi_{P,Q}^2):=\{(p,q^2)\}$, where each contains a single distribution pair on $(\mathcal{X}\times\mathcal{X})\subseteq \mathbb{R}^2\times \mathbb{R}^2$. The source distribution $p$, shared by both $\pi^1$ and $\pi^2$, is uniformly distributed over the arc $\{(\cos\theta^{\circ},\sin\theta^{\circ}): \theta\in [-90,0]\}$. The target distributions differ: $q^1$ is uniform over $\{(\cos\theta^{\circ},\sin\theta^{\circ}): \theta\in [90,180]\}$, and $q^2$ over $\{(\cos\theta^{\circ},\sin\theta^{\circ}): \theta\in [0,90]\}$, with all distributions supported on arcs of the unit circle centered at the origin.  The function distributions $\pi^1_{F\vert P,Q}(\cdot\vert p,q^1)$ and $\pi^2_{F\vert P,Q}(\cdot\vert p,q^2)$ are uniform over the hypothesis class $\mathcal{H}:=\{f^c:c\in[0,360]\}$, where $f^c$ denotes a linear separator rotated by $c^{\circ}$ clockwise from the upward vertical axis and its clockwise side is labelled as class $1$. Figure \ref{Fig:eg2} provides a geometric illustration of the distribution supports and example classifiers.

As computed in Appendix \ref{apx:comp_examples}, the optimal overall target domain risks $R^*_{\infty}$ of two UDA classes are $0$ and $0.125$, respectively, implying that $\pi^1$ is the easiest while $\pi^2$ is relatively challenging. We then select one of the hardest UDA instances from each UDA class: $(p,q^1,f^0)$ from $\pi^1$ and $(p,q^2,f^0)$ from $\pi^2$. In both UDA instances,  since the source distribution and ground-truth function are identical, so the set of sample-consistent classifiers is the same, given by $\{f^c: c\in[0,90]\}$. Moreover, both posteriors $\rho^1(\cdot\vert p,q^1,f_p^0)$ and $\rho^2(\cdot\vert p,q^2,f_p^0)$ are uniform over this set. The corresponding optimal learners are:  $\left((\rho^1)^{\rm A}\right)^{\rm H} (x\vert p,q^1,f^0_p)=1, \forall x\in\Omega(q^1)$ and $\left((\rho^2)^{\rm A}\right)^{\rm H} (x\vert p,q^2,f^0_p)= f^{45}(x), \forall x\in\Omega(q^2)$. Accordingly, the optimal sample-wise risks are $e^{1*}(p,q^1,f^0_p)=0$ and $e^{2*}(p,q^2,f^0_p)=0.25$.

In the UDA instance $(p,q^1,f^0)$, all classifiers $f\in\mathcal{H}$ perform identically on both $p$ and $q^1$  (i.e. $R(f\vert p,f^0)=R(f\vert q^1,f^0), \forall f\in\mathcal{H}$). Therefore, any classifier that performs perfectly on the source domain will also perform perfectly on the target domain, making this UDA instance an easy transfer task. However, in the UDA instance $(p,q^2,f^0)$, a perfect classifier for $p$ may have drastically different performance on $q^2$. As a result, learning a classifier with low target domain risk becomes challenging, making the transfer from source to target relatively harder. 

Our PTLU for two observations are $U^1(p,q^1,f^0_p)=0$ and $U^2(p,q^2,f^0_p)=\int_{0}^{90}\mathscr{H}\left({\rm Ber}(x/90)\right)/90 dx=0.5$, accurately reflecting the transfer difficulty, with the first instance being easiest and the second relatively harder. As shown in Table \ref{tab:values of examples}, we also observe that $\mathcal{H}\Delta\mathcal{H}$ divergence, $\mathcal{Y}$-divergence, and marginal transfer exponent also successfully capture this transfer difficulty, while some traditional divergence measures fail to capture this difficulty. In particular,  $f$-divergence\textemdash{}gives rise to an infinity value, due to $p \centernot{\ll} q^1$ and $p \centernot{\ll} q^2$ (i.e., $p$ is not absolutely continuous with respect to 
$q^1$ or $q^2$) and suggests wrongly the impossibility of transfer in both instances. Similarly, the Wasserstein distance, an Integral Probability Metric (IPM), between source and target distributions is larger in the first instance than in the second, incorrectly implying greater transfer difficulty in the first case. These results demonstrate that some traditional metrics and divergences cannot directly reflect transfer difficulty from source to target domains.

\textbf{Example 2.} Similar to Example 1, we define two UDA classes, $\pi^1$ and $\pi^2$, bellow. While they are structurally ``symmetric" (i.e., one involves transfer from $p$ to $q$, and the other from $q$ to $p$), their transfer difficulty are ``asymmetric".  But,  measures relying on the absolute value operator fail to capture this asymmetry.

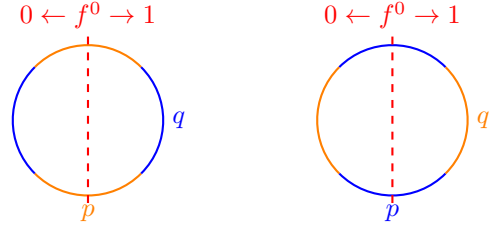
\begin{wrapfigure}{R}{0.5\textwidth}
    \begin{minipage}{0.24\textwidth}
    \centering
    \begin{tikzpicture}
        \draw[thick, orange] (-0.7071,0.7071) arc[start angle=135, end angle=45, radius=1cm];
        \draw[thick, orange] (0.7071,-0.7071) arc[start angle=-45, end angle=-135, radius=1cm];
        \draw[thick, blue] (0.7071,0.7071) arc[start angle=45, end angle=-45, radius=1cm];
        \draw[thick, blue] (-0.7071,-0.7071) arc[start angle=-135, end angle=-225, radius=1cm];

        \draw[red, thick, dashed] (0,-1.1) -- (0,1.2);
        
        \node[below, orange] at (0,-1) {$p$};
        \node[right, blue] at (1,0) {$q$};
        \node[above, red] at (0,1.1) {$0 \leftarrow f^0 \rightarrow 1$};

        \end{tikzpicture}
    \end{minipage}
    \begin{minipage}{0.24\textwidth}
    \centering
    \begin{tikzpicture}
        \draw[thick, blue] (-0.7071,0.7071) arc[start angle=135, end angle=45, radius=1cm];
        \draw[thick, blue] (0.7071,-0.7071) arc[start angle=-45, end angle=-135, radius=1cm];
        \draw[thick, orange] (0.7071,0.7071) arc[start angle=45, end angle=-45, radius=1cm];
        \draw[thick, orange] (-0.7071,-0.7071) arc[start angle=-135, end angle=-225, radius=1cm];

        \draw[red, thick, dashed] (0,-1.1) -- (0,1.2);
        
        \node[below, blue] at (0,-1) {$p$};
        \node[right, orange] at (1,0) {$q$};
        \node[above, red] at (0,1.1) {$0 \leftarrow f^0 \rightarrow 1$};

        \end{tikzpicture}
    \end{minipage}
   \caption{Distributions $p,q$ and the classifier $f^0$ of two UDA classes $\pi^1$ (left) and $\pi^2$ (right) in Example 2.}
   \label{Fig:eg3}
\end{wrapfigure}

Let $\Omega(\pi_{P,Q}^1):=\{(p,q)\}$ and  $\Omega(\pi_{P,Q}^2):=\{(q,p)\}$, where $p$ is uniformly distributed over the arcs $\{(\cos\theta^{\circ},\sin\theta^{\circ}): \theta\in [-45,45]\cup[135,225]\}$ and $q$ over $\{(\cos\theta^{\circ},\sin\theta^{\circ}): \theta\in [45,135]\cup[225,315]\}$, as shown in Figure \ref{Fig:eg3}. The function distributions $\pi^1_{F\vert P,Q}(\cdot\vert p,q)$ and $\pi^2_{F\vert P,Q}(\cdot\vert q,p)$ are both uniformly distributed over $\mathcal{H}:=\{f^c:c\in[-45,45]\}$, where $f^c$ is defined as in Example 1. 

As computed in Appendix \ref{apx:comp_examples}, we have $R_{\infty}^{1*}=0$ for $\pi^1$ while $R_{\infty}^{2*}=0.25$ for $\pi^2$, indicating that $\pi^2$ is harder. We then select one of the hardest UDA instances, $(p,q,f^0)$ and $(q,p,f^0)$, from $\pi^1$ and $\pi^2$, respectively. In the UDA instance $(p,q,f^0)$, only $f^0$ achieves zero error on the source domain, resulting in $\Omega\left(\rho^1(\cdot\vert p,q,f_p^0)\right)=\{f^0\}$, which leads to a PTLU value of $0$. But, in the second instance, the posterior $\rho^2(\cdot\vert q,p,f_p^0)$ is uniform over $\mathcal{H}$, yielding a PTLU value of $0.5$. In both instances, the optimal learner is the same: $\left((\rho^1)^{\rm A}\right)^{\rm H} (x\vert p,q,f^0_p)=\left((\rho^2)^{\rm A}\right)^{\rm H} (x\vert q,p,f^0_p)=f^0(x)$ for $x$ on the support of their target distributions.  The sample-wise risks  of the optimal learner are $e^{1*}(p,q,f^0_p)=0$ and $e^{2*}(q,p,f^0_p)=0.25$.
 
These PTLU values highlight the asymmetry in transfer: the difficulty of  transferring from $p$ to $q$ may differ from that of transferring from $q$ to $p$, even under the same ground-truth function and function class. Specifically, the first UDA instance is an easy task since any classifier $f\in\mathcal{H}$ with a source domain risk below $1/2$ will achieve zero target risk (i.e., $R(f\vert p,f^0)\leq 1/2 \Rightarrow R(f\vert q,f^0)=0$ ). In contrast, in the second UDA instance, even if the learner obtains a classifier with zero source domain error, its target risk is likely to be substantially high. Marginal transfer exponent correctly reflects such transfer difficulty too. However, in both UDA instances, the $f$-divergence is still infinite, the Wasserstein distances are the same, while the $\mathcal{H}\Delta\mathcal{H}$ divergence and $\mathcal{Y}$-discrepancy are both $1/2$. Most measures that incorporate the absolute value operator in their formulations fail to capture the asymmetry in transfer difficulty.

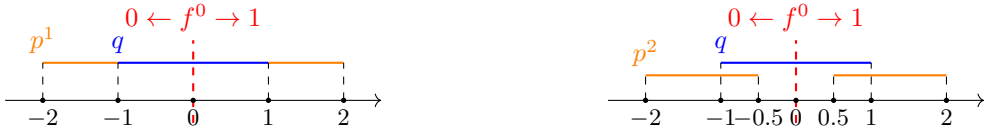
\begin{figure}[h]
    \begin{minipage}{0.48\textwidth}
        \centering 
        \begin{tikzpicture}
            \draw[->] (-2.5, 0) -- (2.5, 0) ; 
            \draw[thick, red, dashed] (0, -0.3) -- (0, 0.8); 
            \node[above, red] at (0,0.8) {$0 \leftarrow f^0 \rightarrow 1$}; 
            \fill[black] ( -2,0) circle (1pt); 
            \fill[black] (-1,0) circle (1pt); 
            \fill[black] (0, 0) circle (1pt); 
            \fill[black] (1,0) circle (1pt); 
            \fill[black] (2,0) circle (1pt); 
            \node[below] at ( -2,0)  {\small $-2$}; 
            \node[below] at ( -1,0) {\small $-1$}; 
            \node[below] at (0, 0) {\small $0$}; 
            \node[below] at ( 2,0)  {\small $2$}; 
            \node[below] at ( 1,0) {\small $1$}; 
    
            \draw[thick, blue] (-1, 0.5) -- (1, 0.5);  
            \draw[thick, orange] (1, 0.5) -- (2, 0.5);  
            \draw[thick, orange] (-2, 0.5) -- (-1, 0.5);  
            
            \draw[dashed] (-2, 0) -- (-2,0.5);    
            \draw[dashed] (-1, 0) -- (-1,0.5);    
            \draw[dashed] (2, 0) -- (2,0.5);    
            \draw[dashed] (1, 0) -- (1,0.5);  
            \node[above, orange] at (-2,0.5) {$p^1$}; 
            \node[above, blue] at (-1,0.5) {$q$}; 
        \end{tikzpicture}
        
        \end{minipage}
    \begin{minipage}{0.48\textwidth}
        \centering 
        \begin{tikzpicture}
            \draw[->] (-2.5, 0) -- (2.5, 0) ; 
            \draw[thick, red, dashed] (0, -0.3) -- (0, 0.8); 
            \node[above, red] at (0,0.8) {$0 \leftarrow f^0 \rightarrow 1$}; 
            \fill[black] ( -2,0) circle (1pt); 
            \fill[black] (-1,0) circle (1pt); 
            \fill[black] (0, 0) circle (1pt); 
            \fill[black] (1,0) circle (1pt); 
            \fill[black] (2,0) circle (1pt); 
            \fill[black] (0.5,0) circle (1pt); 
            \fill[black] (-0.5,0) circle (1pt); 
            \node[below] at ( -2,0)  {\small $-2$}; 
            \node[below] at ( -1,0) {\small $-1$}; 
            \node[below] at (0, 0) {\small $0$}; 
            \node[below] at ( 2,0)  {\small $2$}; 
            \node[below] at ( 1,0) {\small $1$}; 
            \node[below] at ( -0.5,0)  {\small $-0.5$}; 
            \node[below] at ( 0.5,0) {\small $0.5$}; 
    
            \draw[thick, blue] (-1, 0.5) -- (1, 0.5);  
            \draw[thick, orange] (1/2, 1/3) -- (2, 1/3);  
            \draw[thick, orange] (-1/2, 1/3) -- (-2, 1/3);  
            
            \draw[dashed] (-2, 0) -- (-2,1/3);    
             \draw[dashed] (-1/2, 0) -- (-1/2,1/3);    
            \draw[dashed] (-1, 0) -- (-1,0.5);    
            \draw[dashed] (2, 0) -- (2,1/3);    
            \draw[dashed] (1/2, 0) -- (1/2,1/3);    
            \draw[dashed] (1, 0) -- (1,0.5);    
            \node[above, orange] at (-2,0.33) {$p^2$}; 
            \node[above, blue] at (-1,0.5) {$q$}; 
        \end{tikzpicture}
        \end{minipage}  
    \caption{Visualization of distributions $p^1,p^2,q$ and the classifier $f^0$  of two UDA instances in Example 3.}
    \label{Fig:eg4}
\end{figure}
\textbf{Example 3.} 
Again, we define two UDA classes, $\pi^1$ and $\pi^2$, with the latter being more difficult. In this example, we show that the novel measure\textemdash{}transfer exponent\textemdash{}fails to distinguish between these two UDA classes.  

Let $\Omega(\pi_{P,Q}^1):=\{(p^1,q)\}$ and  $\Omega(\pi_{P,Q}^2):=\{(p^2,q)\}$, where $p^1$ is uniformly distributed over intervals $[-2,-1]\cup [1,2]$, $p^2$ over $[-2,-1/2]\cup [1/2,2]$, and $q$ over $[-1,1]$. Let  both $\pi^1_{F\vert P,Q}(\cdot\vert p^1,q)$ and $\pi^2_{F\vert P,Q}(\cdot\vert p^2,q)$ be uniform on $\mathcal{H}:=\{f^c:c\in[-1,1]\}$, where $f^c(x):=\mathbbm{1}\{x\geq c\}$ represents the linear separator for $x\in\mathbb{R}$, as shown in Figure \ref{Fig:eg4}.

Similarly, the overall target domain risks for the optimal learner are $R_{\infty}^{1*}=0.25$ for the first UDA class and $R_{\infty}^{2*}\approx 0.06$ for the second, indicating that $\pi^1$ is harder. Next, we choose the hardest UDA instances $(p^1,q,f^0)$ and $(p^2,q,f^0)$ from two classes respectively. Then, the posterior $\rho^1(\cdot\vert p^1,q,f^0_{p^1})$ is uniform on $\{f^c:c\in[-1,1]\}$ but $\rho^2(\cdot\vert p^2,q,f^0_{p^2})$ is uniform on the subset $\{f^c:c\in[-1/2,1/2]\}$. This leads to the aggregated distributions:  $(\rho^1)^{\rm A}(\cdot\vert x;p^1,q,f^0_{p^1})=[(1-x)/2,(1+x)/2],\forall x\in[-1,1]$ and $(\rho^2)^{\rm A} (\cdot\vert x;p^2,q,f^0_{p^2})=[1/2-x,1/2+x],\forall x\in [-1/2,1/2]$. In both instances, the optimal learner outputs the ground-truth classifier: $\left((\rho^1)^{\rm A}\right)^{\rm H} (x\vert p^1,q,f^0_{p^1})=\left((\rho^2)^{\rm A}\right)^{\rm H} (x\vert p^2,q,f^0_{p^2})=f^0(x)$ for any $x$ in the corresponding target support.  The 
the sample-wise risks  of the optimal learner are $e^{1*}(p^1,q,f^0_{p^1})=0.25$ and $e^{2*}(p^2,q,f^0_{p^2})\approx0.05$.
As shown in Appendix \ref{apx:comp_examples}, the PTLU values of these two UDA instances are $0.72$ and $0.36$, respectively, indicating that the second instance is relatively easier.  

The values of $\mathcal{H}\Delta\mathcal{H}$ divergence and $\mathcal{Y}$-divergence in the second instance are lower than those in the first. However, in both UDA instances, the $f$-divergence and the marginal transfer exponent $\gamma$ are both infinite, failing to differentiate the transfer difficulty between these two similar yet distinct UDA instances. In reality, the second instance is relatively easier, as the overlapping regions between $\Omega(p^2)$ and $\Omega(q)$ provide useful classification information for the target domain. Moreover, the expected target domain risk of classifiers with zero source domain error is $1/4$ in the second instance (i.e., $\mathbb{E}_{\rho^2(\cdot\vert p^2,q,f_{p^2}^0)}[R(F\vert q,f^0)]=1/4$), whereas it is $1/2$ in the first instance (i.e., $\mathbb{E}_{\rho^1(\cdot\vert p^1,q,f^0_{p^1} )}[R(F\vert q,f^0)]=1/2$), implying that it is easier to  obtain a classifier with low target domain risk in the second instance.

\textbf{Example 4.}  Now, we consider a single UDA class $\pi$ where $\Omega(\pi_P)=\{p\}$ and $\Omega(\pi_{Q,F})=\{(q^c,f^c),(q^c,\bar{f}^c):c\in \mathbb{R}\}$. As we will show, all the comparison measures fail to reflect the difficulty of $\pi$. 

\begin{wrapfigure}{R}{0.5\textwidth}
\centering 
    \begin{tikzpicture}
        \draw[->] (-1.2, 0) -- (2, 0) node[right] {}; 
        \draw[->] (0, -1.2) -- (0, 2) node[above] {}; 
        \draw[thick, orange] (0, -1) -- (0, 1); 
        \fill[orange] (0, -1) circle (1pt); 
        \fill[orange] (0, 1) circle (1pt); 
        \fill[orange] (0, 0) circle (1pt); 
        \node[left] at (0, 1) {\small $1$}; 
        \node[left] at (0, -1) {\small $-1$}; 
        \node[left] at (0, 0) {\small $0$}; 
        \node[right, orange] at (0, 1.1) {$p$}; 
        \draw[thick, blue] (1, 0.5) -- (1, 2.5); 
        \fill[blue] (1, 0.5) circle (1pt); 
        \fill[blue] (1, 2.5) circle (1pt); 
        \fill[blue] (1, 1.5) circle (1pt); 
        \node[left] at (1, 2.5) {\small $c+1$}; 
        \node[left] at (1, 1.5) {\small $c$}; 
        \node[left] at (1, 0.5) {\small $c-1$}; 
        \node[right, blue] at (1, 2.5) { $q^c$}; 
        \draw[thick, red, dashed] (-0.5,-0.75) -- (1.5, 2.25); 
        \node[right, red] at (1.5, 2.25) {$f^c$}; 
        \node[red,above ] at (1.2, 2.1) { $1$}; 
        \node[red,below ] at (1.5, 1.9) {$0$};
    \end{tikzpicture}
\caption{Visualization of distributions $p,q^c$ and classifiers $f^c$ in Example 4.}
    \label{Fig:eg1}
\end{wrapfigure}
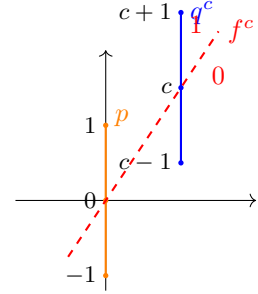

As shown in Figure \ref{Fig:eg1},  let $p$ be a uniform distribution on the line segment connecting points $(0,-1)$ and $(0,1)$, $q^c$ be uniform on the line segment between $(1,c-1)$ and $(1,c+1)$. The labeling functions $f^c$ and $\bar{f}^c$ are defined as:   

\[
         f^c(x):=1-\bar{f}^c(x):=\mathbbm{1}\left\{x^T \begin{bmatrix}
        0 & -1 \\
        1 &  0\\
        \end{bmatrix} \begin{bmatrix}
        1 \\
        c \\
        \end{bmatrix} \geq 0\right\}.
\]
    
Let $\mathcal{H}:=\{f^c,\bar{f}^c:c\in \mathbb{R}\}$. Then, the overall target domain risk of the optimal learner is $0$, implying that  this UDA class has the lowest transfer difficulty. We then draw one of the hardest UDA instances, $(p,q^{c^*},f^{c^*})$, from $\pi$ with any parameter $c^*$. In this instance, we can directly compute $\Omega\left(\rho(\cdot\vert p,q^{c^*},f_p^{c^*})\right)=\{f^{c^*}\}$, leading to $\rho^{\rm A}(\cdot\vert x;p,q^{c^*},f_p^{c^*})$ being one-hot for any $x\in\Omega(q^{c^*})$ and the PTLU being $0$. This result arises because the direction of classifiers ($f$ or $\bar{f}$) can be uniquely determined using the labeled source sample, and the value of parameter can be identified from the unlabeled target sample. Besides, the optimal learner in this UDA instance is $\left(\rho^{\rm A}\right)^{\rm H}(x\vert p,q^{c^*},f_p^{c^*})=f^{c^*}(x), \forall x\in\Omega(q^{c^*})$ and its sample-wise risk  is $e^*(p,q^{c^*},f^{c^*}_p)=0$.

However, the $f$-divergence between $p$ and $q^{c^*}$ is infinity, the $\mathcal{H}\Delta\mathcal{H}$ divergence for this example is $1$, the $\mathcal{Y}$-discrepancy is $1/2$, and the transfer exponent $\gamma$ is infinity,(as there exists $h\in\mathcal{H}$ such that $\Pr_{ p} (h(X)\neq f(X))>0$ while $\Pr_{q^{c^*}} (h(X)\neq f(X))=0$). These metrics suggest that this UDA instance is challenging or even impossible.  But, transferring from $p$ to $q^{c^*}$ is straightforward in practice: We require only $1$ labeled source data that is not located at $(0,0)$ to determine the classifier's direction. The accuracy in determining the value of $c$ depends on the size of target sample, which can be computed via Hoeffding inequality.



\subsection{Risk Lower Bounds from Empirical Posterior Target Label Uncertainty}

So far, we see that both the notions of $e^*(s_{m, n}, q)$ in Theorem \ref{thm:optimal_mn} and the  PTLU based lower bound in Theorem \ref{thm:LB_mn}  may be used to characterize the difficulty of learning a UDA instance $(p, q, f)$ from its sample, the computation of these terms requires the precise knowledge of the target domain distribution $q$. Next we provide a lower bound that is based on ``empirical PTLU", which only depends on the target sample ${\bf x}_{\rm t}$ and requires no additional konwledge of $q$. Specifically, the {\em Empirical Posterior Target Label Uncertainty (EPTLU)} on an $(m, n)$-sample $s_{m, n}$ is defined as
\begin{equation}
\widetilde{U}(s_{m, n}) := \mathbb{E}_{x \sim {\bf x}_{\rm t}} \mathscr{H}\left(\rho^{\rm A}(\cdot\vert x, s_{m, n})\right).
\end{equation}

\begin{thm}
Assume that $\mid\mathcal{Y}\mid=k$. For any learner $\mathscr{A}_{m,n}$, and any $t>0$, with probability at least $1-\exp\left(-\dfrac{2nt^2}{(\log k)^2}\right)$ in the sampling process of $s_{m, n}$ which realizes an UDA instance $(p, q, f)$, the following inequalities hold:
    \begin{equation}
    \begin{aligned}
    \label{eq:LB_mn}
       k>2:  e(\mathscr{A}_{m,n}; s_{m, n}, q)&\geq \frac{\widetilde{U}(s_{m,n})-t-1}{\log (k-1)}\\
        k=2:  e(\mathscr{A}_{m,n}; s_{m, n}, q)&\geq \frac{\big(\widetilde{U}(s_{m,n})-t\big)^2}{4}+e^*(s_{m,n}, q)^2
    \end{aligned}
    \end{equation}
\label{thm:finite to infinite wrt xt and q}
\end{thm}

It is also possible to use EPTLU\textemdash{}computed from a finite sample\textemdash{}to estimate the learning difficulty on an infinite-sample observation. To that end, we set up some notations and regularity conditions.

Denote $m\wedge n:=\min\{m,n\}$, and $\mathcal{E}:=\Delta(\mathcal{X})\times\Delta(\mathcal{X})$. 
Let ${\cal M}(\epsilon, {\cal T}, d)$ denote the $\epsilon$-packing number of a set ${\cal T}$ under metric $d$.
We will use $d_{\rm TV}$ to denote the total variation distance
and $h^2$ to denote the square of Hellinger distance. We will consider the case where ${\cal X}$ is finite, and for any distribution $p$ on ${\cal X}$ denote $\alpha_p:=\min_{x\in \Omega(p)} p(x)$, $N_p:=\mid\Omega(p)\mid$. We denote $\beta:=\underset{x\in\Omega(q)}{\min}\underset{y\in\mathcal{Y}}{\min}\;\rho^{\rm A}(y\vert x,p,q,f_p)$. We make the following regularity assumptions.

\paragraph{(A1)} There are a sequence $(\epsilon_i)$ of positive numbers and a sequence $(\mathcal{E}_i)$ of subsets of $\mathcal{E}$, such that 
        $\lim_{i\rightarrow \infty} \epsilon_i=0$, $\lim_{i\rightarrow \infty} i\epsilon_i^2=\infty$, 
        $\lim_{i\rightarrow \infty} 
        \sum\limits_{j=1}^i \exp(-Bj\epsilon_j^2) < \infty$ for all $B>0$, $\log \mathcal{M}(\epsilon_{i},\mathcal{E}_{i},d_{\rm TV})\leq {i}\cdot\epsilon_{i}^2$, 
        and for some $C>0$, $\pi_{PQ}\Big(\mathcal{E}\backslash\mathcal{E}_{i}\Big)\leq \exp\big(-{i}\cdot \epsilon  _{i}^2(C+4)\big)$, $\pi_{PQ}\Big({(p',q'):h^2\big(p'\otimes q',p\otimes q\big)\cdot\bigg\Vert{\dfrac{p'\otimes q'}{p\otimes q}}\bigg\Vert_\infty}\leq \epsilon_{i}^2\Big)\geq \exp\big(-{i}\cdot\epsilon_{i}^2 C\big)$.

\paragraph{(A2)} For any $f'\in\mathcal{F}$, $\pi_{F\vert P,Q}(f'\vert p,q)$,  as a function of $(p, q)$, is $L$-Lipschitz with respect to $d_{\rm TV}$. 



\paragraph{(A3)}$m>N_p$, $n>N_q$, and $\beta>0$.

\begin{thm}
Let $k:=\mid\mathcal{Y}\mid$, $K:=\big\vert\mathcal{F}[f_p]\big\vert$, and $S:=\underset{f'\in\mathcal{F}[f_p]}{\sum}\pi_{F\vert P,Q}(f'\vert p,q)$. Assume (A1), (A2) and (A3) hold. 
For a constant $A=\mathcal{O}\Bigg(\dfrac{LK^3\log({1/\beta})}{S^2}\Bigg)$, with probability at least $1-N_qK^2\exp\left(-{m\wedge n}\cdot\epsilon_{m\wedge n}^2\right)-N_p\exp(-m\alpha_p)-N_q\exp(-n\alpha_q)$ under the sampling of $s_{m,n}$
that realizes a UDA instance $(p, q, f)$,
 any learner $\mathscr{A}_{m,n}$ 
 satisfies
    \begin{equation}
    \begin{aligned}
    \label{eq:LB_mn}
        k>2:  e(\mathscr{A}_{\infty}; p, q, f_p) &\geq \frac{\widetilde{U}(s_{m,n})-A\cdot\epsilon_{m\wedge n}-1}{\log (k-1)}\\
        k=2:  e(\mathscr{A}_{\infty}; p, q, f_p)&\geq \frac{\big(\widetilde{U}(s_{m,n})-A\cdot\epsilon_{m\wedge n}\big)^2}{4}+e^*(s_{m,n}, q)^2
    \end{aligned}
    \end{equation}
 \label{thm:finite to infinite}
 \end{thm}

From the theorem, the convergence of the probability
${\rm Pr}[(\ref{eq:LB_mn}) ~{\rm fails}]$ to zero is dominated by the term 
$\exp\left(-{m\wedge n}\cdot\epsilon_{m\wedge n}^2\right)$. If $\epsilon_i$ is chosen to decay as $i^{-\kappa}$ for some $\kappa\in (0, 1/2)$, the term decays to zero as $\exp\left(
-(m\wedge n)^{1-2\kappa} 
\right)$. 

\section{Concluding Remarks}
This paper theoretically studies the hardness of learning in Unsupervised Domain Adaptation. Although the key contributions of this work are of theoretical nature, we now remark the practical relevance of this work.

In practice, when we seek to find a classifier in a ``target domain'', we may decide to use a UDA approach by exploiting some labeled data in a related source domain. Under the framework presented in this paper, the foremost step is to decide $\pi$, the prior distribution of the ground-truth triple $(p, q, f)$, and this distribution should reflect the practioner's knowledge on how the source distribution $p$, the target distribution $q$ and the ground-truth classifier $f$ are linked.  The results of this paper suggest that such knowledge governs the fundamental difficulty of learning and that this difficulty can be estimated using PLTU. These results then potentially provide guidelines for the practitioners, for example, allowing one to estimate if the choice of source domain is appropriate or to decide if one choice is better than another.\textemdash{}One should only consider the UDA approach if the resulting learning task is not too difficult.

The practical learning scenarios are however often challenged by lacking precise specification of $\pi$. This makes it difficult to fully exploit the development in this paper. In this case, the guideline provided by our results may still take effect,  at least at a high level. On the other hand, we believe that there are real-world scenarios in which practitioners can make precise their prior knowledge of $\pi$, for example, when $p$ and $q$ are described by related graphical models (such as two Bayesian networks sharing a common structure). We anticipate that exploration in this direction lead to interesting and useful discoveries. 

\clearpage
\bibliography{src/collas2025_conference}
\bibliographystyle{src/collas2025_conference}

\appendix
\newpage
\section*{Appendix}
\section{Summary of Notations}
\label{apx: notations}
We provide a table summarizing the notations used in this work, as shown in Table \ref{tab:notations}.

\begin{table}[h]
\caption{Notations in this work}
    \begin{tabular}{c|l}
    \hline
    Notations  & Descriptions\\
    \hline\hline
        $\mathcal{X,Y}$ & Input and output spaces \\
        $\mathcal{F}$ & The function space: a set of measurable functions mapping $\mathcal{X}$ to $\mathcal{Y}$\\
        $\Delta(\mathcal{X})$, $\Delta(\mathcal{F})$ &  The sets of all probability distributions on $\mathcal{X}$ and $\mathcal{F}$, respectively\\ 
        $\Omega(p)$   & The support of a distribution $p$\\
       $\pi$  &  A UDA class: a distribution over $\Delta(\mathcal{X}) \times \Delta(\mathcal{X}) \times \mathcal{F}$\\
      $(p,q,f)\sim\pi$   & A UDA instance: a triplet sampled from $\pi$ \\
      \hline
      $m$, $n$ & The sizes of source and target samples, respectively \\
       $s_{m,n}\sim (p,q,f)$  & An observed sample drawn from $(p, q, f)$\\
        $\mathcal{S}_{m,n}$ & The sample space: $\mathcal{X}^m\times\mathcal{X}^n\times\mathcal{Y}^m$\\
        $\mathscr{A}_{m,n}$ & A learner input a sample in $\mathcal{S}_{m,n}$ to a distribution over $\mathcal{F}$\\
        $\mathscr{A}_{m,n}(\cdot\vert s_{m,n})$ & The distribution over $\mathcal{F}$ returned by $\mathscr{A}_{m,n}$ after observing $s_{m,n}$\\
        $\mathscr{A}^*_{m,n}$ & The optimal learner\\
        \hline
        $R(g|q,f)$ & The target domain risk of a single classifier $g\in\mathcal{F}$ w.r.t. $f$ on $q$\\
        $R(\mathscr{A}_{m,n}|p,q,f)$ & The expected target domain risk of $\mathscr{A}_{m,n}$ w.r.t.  $(p,q,f)$\\
        $R(\mathscr{A}_{m,n})$ & The overall target domain risk of  $\mathscr{A}_{m,n}$\\
        $e(\mathscr{A}_{m,n};s_{m,n},q)$ & The sample-wise risk of $\mathscr{A}_{m,n}$ on $q$ when observing $s_{m,n}$\\
        $e^*(s_{m,n},q)$ & The sample-wise risk of $\mathscr{A}^*_{m,n}$ on $q$ when observing $s_{m,n}$\\
        $R^*_{m,n}$& The overall target domain risk of $\mathscr{A}^*_{m,n}$\\
        \hline
        $\mathcal{F}[\mathbf{x}_{\rm s},\mathbf{y}_{\rm s}]$ & A subset of functions in $\mathcal{F}$ that label $\mathbf{x}_{\rm s}$ as $\mathbf{y}_{\rm s}$\\
        $f\in\mathcal{F}$ & A hard classifier mapping $\mathcal{X}$ to $\mathcal{Y}$\\
        $f_p$ & A classifier obtained by restricting $f$ from domain $\mathcal{X}$ to domain  $\Omega(p)$\\
        $\mathcal{F}[f_p]$ & A subset of functions in $\mathcal{F}$ that label all $x\in\Omega(p)$ as $f_p(x)$\\
        \hline
        $P,Q$ & The random variables of source and target distributions\\
        $F$ & The random variable of the ground-truth classifier\\
        $X,Y$ & The random variables of the input and its corresponding label\\
        $S$ & The random variable of the observed sample\\
        \hline
        $\rho(\cdot|s_{m,n})$, $\pi_{F|S}(\cdot|s_{m,n})$ & The conditional distribution of function $F$ given a sample $S=s_{m,n}$\\
        $\rho^{\rm A}(\cdot|s_{m,n})$ & The aggregated classifier of distribution $\rho(\cdot|s_{m,n})$\\
        $\left(\rho^{\rm A}\right)^{\rm H}(\cdot|s_{m,n})$ & The hardened version of soft classifier $\rho^{\rm A}(\cdot|s_{m,n})$\\
    \hline  
    $\mathscr{H}(\cdot)$ & The entropy functional of a given distribution\\
    $H(\cdot)$ & The entropy of a given random variable\\
    $U(\cdot)$ & The PTLU on a given sample\\
    \hline  
    \end{tabular}
    \label{tab:notations}
\end{table}

\section{Decomposition of the Overall Target Domain Risk}
\label{apx:decp}
\paragraph{1. $m,n$-version:} Fix the input point $x$, then for any sample $s_{m,n}$, we define $R(\mathscr{A}_{m,n},x,s_{m,n})$ as the overall risk of $\mathscr{A}_{m,n}$ at point $x$:
\[
R(\mathscr{A}_{m,n},x,s_{m,n}) :=\mathbb{E}_{f\sim\rho(\cdot|s_{m,n})}\mathbb{E}_{g\sim\mathscr{A}_{m,n}(\cdot|s_{m,n})} \mathbbm{1}\{g(x)\neq f(x)\}
\]
Taking expectation over $s_{m,n}\sim (p,q,f)$ and $x\sim q$ on $R(\mathscr{A}_{m,n},x,s_{m,n})$ , by the Markov property, it becomes the expected target domain risk of $\mathscr{A}_{m,n}$:
\[
R(\mathscr{A}_{m,n}|p,q,f)=\mathbb{E}_{s_{m,n}\sim(p,q,f) }\mathbb{E}_{x\sim q }[R(\mathscr{A}_{m,n},x,s_{m,n}) ]
\]

Then, the over all target domain risk of $\mathscr{A}_{m,n}$ is
\begin{equation}
\begin{aligned}
\label{eq:dcp_mn}
R(\mathscr{A}_{m,n})&=\mathbb{E}_{(p,q,f)\sim \pi}[R(\mathscr{A}_{m,n}|p,q,f)]\\
&=\mathbb{E}_{(p,q,f)\sim \pi}\mathbb{E}_{s_{m,n}\sim(p,q,f) }\mathbb{E}_{f\sim\rho(\cdot|s_{m,n})}\mathbb{E}_{g\sim\mathscr{A}_{m,n}(\cdot|s_{m,n})}\mathbb{E}_{x\sim q }[\mathbbm{1}\{g(x)\neq f(x)\}]\\
&=\mathbb{E}_{(p,q,f)\sim \pi}\mathbb{E}_{s_{m,n}\sim(p,q,f) }[e(\mathscr{A}_{m,n};s_{m,n},q)]
\end{aligned}
\end{equation}
\paragraph{2. $\infty$-version:} By definition, the overall target domain risk of $\mathscr{A}_{\infty}$ can be written as
\begin{align*}
    R(\mathscr{A}_{\infty})&=\mathbb{E}_{(p,q,f)\sim \pi}\mathbb{E}_{g\sim\mathscr{A}_{\infty}(\cdot|p,q,f_p)}\mathbb{E}_{x\sim q} \mathbbm{1}\{g(x)\neq f(x)\}\\
    &=\mathbb{E}_{(p,q,f,f_p,g,x)\sim \nu} \mathbbm{1}\{g(x)\neq f(x)\}\\
\end{align*}
where $\nu$ represents the joint distribution of these random variables. According to the Markov chain they form, we can reorder the expectations as follows:
\begin{equation}
\begin{aligned}
\label{eq:dcp_infty}
    R(\mathscr{A}_{\infty})&=\mathbb{E}_{(p,q,f_p)\sim \nu_{P,Q,F_P}}\mathbb{E}_{(p,q,f)\sim\nu_{P,Q,F|P,Q,F_P}(\cdot,\cdot,\cdot|p,q,f_p)}\mathbb{E}_{g\sim\nu_{G|P,Q,F_P}(\cdot|p,q,f_p)} \mathbb{E}_{x\sim \nu_{X|Q}(\cdot|q)}\mathbbm{1}\{g(x)\neq f(x)\}\\
    &=\mathbb{E}_{(p,q,f_p)\sim \pi_{P,Q,F_P}}\mathbb{E}_{f\sim\rho(\cdot|p,q,f_p)}\mathbb{E}_{g\sim\mathscr{A}_{\infty}(\cdot|p,q,f_p)} \mathbb{E}_{x\sim q}\mathbbm{1}\{g(x)\neq f(x)\}\\
    &=\mathbb{E}_{(p,q,f)\sim \pi}\mathbb{E}_{f'\sim\rho(\cdot|p,q,f_p)}\mathbb{E}_{g\sim\mathscr{A}_{\infty}(\cdot|p,q,f_p)} \mathbb{E}_{x\sim q}\mathbbm{1}\{g(x)\neq f'(x)\}\\
    &=\mathbb{E}_{(p,q,f)\sim \pi}[e(\mathscr{A}_{\infty};p,q,f_p)]
\end{aligned}
\end{equation}
It is easy to show Eq. \ref{eq:dcp_mn} is equivalent to Eq. \ref{eq:dcp_infty} as $m$ and $n$ tend to infinity.

\section{Proof of Optimal Learners}
\label{apx:optimal}
For any UDA instance $(p,q,f)\sim\pi$ and any sample $s_{m,n}\sim (p,q,f)$, we show the optimality of $\left(\rho^{\rm A}\right)^{\rm H}(\cdot|s_{m,n})$ and $\left(\rho^{\rm A}\right)^{\rm H}(\cdot|p,q,f_p)$ separately:
\paragraph{1. $m,n$-version:}  For any learner $\mathscr{A}_{m,n}$, we have:
\begin{align*}
e(\mathscr{A}_{m,n};s_{m,n},q)
&=1-\mathbb{E}_{x\sim q }\mathbb{E}_{g\sim\mathscr{A}_{m,n}(\cdot|s_{m,n})}\left[\Pr_{\rho(\cdot|s_{m,n})}(g(x)= F(x))\right]\\
&=1-\mathbb{E}_{x\sim q }\mathbb{E}_{g\sim\mathscr{A}_{m,n}(\cdot|s_{m,n})}\left[\rho^{\rm A}(g(x)|x,s_{m,n})\right]\\
e(\mathscr{A}^*_{m,n};s_{m,n},q)
&=1-\mathbb{E}_{x\sim q }\left[\max_{y}\rho^{\rm A}(y|x,s_{m,n})\right]
\end{align*}
Thus,
\[
e(\mathscr{A}^*_{m,n};s_{m,n},q)\leq e(\mathscr{A}_{m,n};s_{m,n},q)
\]
which proves the optimality of $\mathscr{A}^*_{m,n}$. This result can be easily extend to the case where $m,n\rightarrow\infty,\infty$. We can also prove it using the definition of $e(\mathscr{A}_{\infty};p,q,f_p)$.

\paragraph{2. $\infty$-version:} For any learner $\mathscr{A}_{\infty}$, we have:
\begin{align*}
    e(\mathscr{A}_{\infty};p,q,f_p)
    &=\mathbb{E}_{f\sim\rho(\cdot|p,q,f_p)}\mathbb{E}_{g\sim\mathscr{A}_{\infty}(\cdot|p,q,f_p)} \mathbb{E}_{x\sim q}\mathbbm{1}\{g(x)\neq f(x)\}\\
    &=1-\mathbb{E}_{x\sim q}\mathbb{E}_{g\sim\mathscr{A}_{\infty}(\cdot|p,q,f_p)} \left[\Pr_{\rho(\cdot|p,q,f_p)}(g(x)= F(x))\right]\\
&=1-\mathbb{E}_{x\sim q}\mathbb{E}_{g\sim\mathscr{A}_{\infty}(\cdot|p,q,f_p)} \left[\rho^{\rm A}(g(x)|x,p,q,f_p)\right]\\
e(\mathscr{A}_{\infty}^*;p,q,f_p)&=1-\mathbb{E}_{x\sim q}\left[\max_{y}\rho^{\rm A}(y|x,p,q,f_p)\right]\\
\end{align*}
Similarly, we have 
\[
e(\mathscr{A}^*_{\infty};p,q,f_p)\leq e(\mathscr{A}_{\infty};p,q,f_p).
\]
$\hfill\square$

 \section{Proof of Thm. \ref{thm:LB_mn}}
\label{apx:proof_LBs}
Unlike the notation $\mathscr{H}(\cdot)$, which denotes the entropy functional, we adopt $H(\cdot)$ to represent the entropy of a random variable and $H_{\rm b}(\cdot)$ to denote the entropy of a Bernoulli random variable with the input as its success probability, e.g.,
\[
H_{\rm b}(c)=c\log\frac{1}{c} +(1-c)\log\frac{1}{1-c}, \forall c\in[0,1].
\]

Let $F$ be a classifier drawn from $\rho(\cdot|s_{m, n})$ and $G$ be a classifier drawn from $\mathscr{A}_{m,n}(\cdot|s_{m,n})$.
For any $X\sim q$, let $Y:=F(X)$ and $\hat{Y}:=G(X)$. Then  $Y$  follows $\rho^{\rm A}(\cdot|X, s_{m,n})$, and $\hat{Y}$ follows $\mathscr{A}_{m,n}^{\rm A}(\cdot|X, s_{m,n})$ and random variables $X,Y,\hat{Y}$ and $S$ form a Markov chain $Y- (X, S) - \hat{Y}$. Then, we can use Fano's inequality to lower bound $e(\mathscr{A}_{m,n};s_{m,n},q)=\Pr(Y\neq \hat{Y};s_{m,n})$ by the conditional entropy $H(Y|X;s_{m,n},q)$, and $U(s_{m,n},q)$ is equivalent to this conditional entropy.

For any given sample $s_{m,n}$ and target distribution $q$, we now prove the upper bound on $e(\mathscr{A}_{m,n};s_{m,n},q)$ by Fano's inequality. Let $E(X|S=s_{m,n}):=\mathbbm{1}\{\hat{F}(X)\neq F(X)|S=s_{m,n}\}=\mathbbm{1}\{\hat{Y}\neq Y|S=s_{m,n}\}$ represent the error indicator on $X$. Using the chain rule for joint entropy, we decompose $H(E(X|S),Y \vert  X,s_{m,n})$ in two different ways:
\begin{equation}
\begin{aligned}
\label{eq:decomp_entropy}
H(E(X|S),Y\vert
\hat{Y},X,s_{m,n})&=\underbrace{H(Y\vert \hat{Y},X,s_{m,n})}_{=H(Y\vert X,s_{m,n})}+\underbrace{H(E(X|S)|\hat{Y},Y,X,s_{m,n})}_{=0}\\
&=H (E(X|S)\vert\hat{Y},X,s_{m,n})+H(Y\vert E(X),\hat{Y},X,s_{m,n})
\end{aligned}
\end{equation}
The conditional independence of these random variables implies that $H(Y|\hat{Y},X,s_{m,n})=H(Y|X,s_{m,n})$. And when $\hat{Y}, Y$ and $X$ are known, there is no randomness in $E(X|s_{m,n})$, resulting in $H(E(X|S)|\hat{Y},Y,X,s_{m,n})=0$. Besides, since $E(X)\not\!\perp\!\!\!\perp \hat{Y}|X$, we have
\begin{align*}
        H(E(X|S)|\hat{Y},X,s_{m,n})&\leq H(E(X|S)|X,s_{m,n})\\
        &=\mathbb{E}_{x\sim q}\left[H(E(X|S)|X=x,s_{m,n})\right]\\
        &\leq H_{\rm b}(\mathbb{E}_{x\sim q}[E(X|S)|X=x,s_{m,n}])\\
        &= H_{\rm b}(e(\mathscr{A}_{m,n};s_{m,n},q))
\end{align*}
where the last inequality holds due to the concavity of entropy function. 

{\bf If $k>2$, }
\begin{align*}
    &\;\;\;\;\;H(Y|E(X|S),\hat{Y},X,s_{m,n})\\
    &=e(\mathscr{A}_{m,n};s_{m,n},q) \cdot H(Y|E(X|S)=1,\hat{Y},X,s_{m,n}) + (1-e(\mathscr{A}_{m,n};s_{m,n},q) )\cdot\underbrace{H(Y|E(X|S)=0,\hat{Y},X,s_{m,n})}_{=0}\\
    &=e(\mathscr{A}_{m,n};s_{m,n},q) \cdot H(Y|E(X|S)=1,\hat{Y},X,s_{m,n}) \leq e(\mathscr{A}_{m,n};s_{m,n},q)\cdot\log(k-1)
\end{align*}

When $E(X|S)=1$ and $\hat{Y}, X$ are known, the possible values of $Y$ are limited to $\mathcal{Y}\backslash\{\hat{Y}\}$. The maximum entropy occurs when the conditional distribution of $Y$ is uniform over the remaining $k-1$ categories,  that is, $H(Y|E(X|X)=1,\hat{Y},X,s_{m,n})\leq \sum_{i=1}^{k-1}\frac{1}{k-1}\log(k-1)=\log(k-1)$.

{\bf If $k=2$,} then $H(Y|E(X|S),\hat{Y},X,s_{m,n})=0$.

Plugging in these inequalities into Eq. \ref{eq:decomp_entropy}, we have 
    \begin{equation}
    \begin{aligned}
              k>2: & H(Y|X,s_{m,s})\leq \underbrace{H_{\rm b}(e(\mathscr{A}_{m,n};s_{m,n},q))}_{\leq 1}+e(\mathscr{A}_{m,n};s_{m,n},q)\cdot\log(k-1)\\
              k=2: & H(Y|X,s_{m,s})\leq H_{\rm b}(e(\mathscr{A}_{m,n};s_{m,n},q)) \leq 2\sqrt{e(\mathscr{A}_{m,n};s_{m,n},q)(1-e(\mathscr{A}_{m,n};s_{m,n},q))}
    \end{aligned}
    \end{equation}
Theorem \ref{thm:LB_mn} is derived by substituting our PTLU, $U(s_{m,n})=H(Y|X,s_{m,n})$.  Corollary \ref{coro:LB_infty} can be proved in the same way or simply by substituting $s_{m,n}\rightarrow (p,q,f_p)$ as $m,n\rightarrow\infty,\infty$.
$\hfill\square$

\section{Proof of Thm. \ref{thm:R(g)}}
\label{apx:proof_lower_bound}
Let $r(g|x,s_{m,n}):= \mathbb{E}_{f\sim \rho(\cdot|s_{m,n})}\mathbbm{1}\{g(x)\neq f(x)\}$ denote the average disagreement between $g$ and all source-consistent classifiers $f$ on a point $x$. Then, we have:
\begin{enumerate}
    \item  $r(g|x,s_{m,n})=1-\rho^{\rm A}(g(x)|x,s_{m,n})$,
    \item $e(\mathscr{A};s_{m,n},q)=\mathbb{E}_{f\sim \rho(\cdot|s_{m,n})}[R(g|q,f)]=\mathbb{E}_{x\sim q}[r(g|x,s_{m,n})]$, where the leaner $\mathscr{A}$ deterministic returns $g$.
\end{enumerate}

When $k>2$, we have:
\begin{align*}
  U(s_{m,n},q)-1& \leq U(s_{m,n},q)-H_{\rm b}(\mathbb{E}_{x\sim q} [r(g|x,s_{m,n})])\\&\leq U(s_{m,n},q)-\mathbb{E}_{x\sim q} [H_{\rm b}(r(g|x,s_{m,n})]\\
    &=\mathbb{E} _{x\sim q}\left[\sum_{y=1}^{k}\left(\rho^{\rm A}(y|x,s_{m,n})\log\frac{1}{\rho^{\rm A}(y|x,s_{m,n})}\right)\right]-\mathbb{E}_{x\sim q} [H_{\rm b}(r(g|x,s_{m,n})(g;x))]\\
    &\leq\mathbb{E} _{x\sim q}\big[r(g|x,s_{m,n})\log\frac{k-1}{r(g|x,s_{m,n})}+(1-r(g|x,s_{m,n}))\log\frac{1}{1-r(g|x,s_{m,n})} \\
    &\;\;\;-r(g|x,s_{m,n})\log\frac{1}{r(g|x,s_{m,n})}-(1-r(g|x,s_{m,n}))\log\frac{1}{1-r(g|x,s_{m,n})} \big]\\
    &=\mathbb{E}_{x\sim q}\left[r(g|x,s_{m,n})\log(k-1)\right]\\
    &= \log(k-1) \mathbb{E}_{f\sim \rho(\cdot|s_{m,n})}[R(g|q,f)]
\end{align*}

When $k=2$, we have:
\begin{align*}
U(s_{m,n},q)&=\mathbb{E}_{x\sim q}\left[H_{\rm b}(r(g|x,s_{m,n}))\right]\\
&\leq H_{\rm b}(\mathbb{E}_{x\sim q}[r(g|x,s_{m,n})])\\
&=H_{\rm b}(\mathbb{E}_{f\sim\rho(\cdot|s_{m,n})}[R(g|q,f)])\\
&\leq 2\sqrt{\mathbb{E}_{f\sim \rho(\cdot|s_{m,n})}[R(g|q,f)]-\mathbb{E}^2_{f\sim \rho(\cdot|s_{m,n})}[R(g|q,f)]}
\end{align*}
The results in Theorem \ref{thm:R(g)} are obtained by applying Chebyshev's inequality.
$\hfill\square$

\section{Computation Processes in Four Examples}
\label{apx:comp_examples}
\paragraph{Example 1.}
The posterior distributions $\rho^1(\cdot|p,q^1,f^c)$ and $\rho^2(\cdot|p,q^2,f^c)$ share the same format:
\begin{align*}
    \rho^1(f^a|p,q^1,f^c_p)=\rho^2(f^a|p,q^2,f^c_p)=\left\{\begin{matrix}
1/90, &  c\in [0,90]\wedge  a\in   [0,90]\\
1/90, &  c\in [180,270]\wedge  a\in   [180,270]\\
1 ,&  c\in [90,180] \cup[270,360] \wedge  a=c\\
0, &  \text{otherwise}
\end{matrix}\right.
\end{align*}
The aggregated versions of  posterior for UDA classes $\pi^1$ and $\pi^2$ are:
\begin{align*}
\forall x\in\Omega(q^1):\left(\rho^1\right)^{\rm A}(y|x,p,q^1,f^c_p)&=\left\{\begin{matrix}
1, &  y = f^c(x) \\
 0, & y\neq f^c(x) \\
\end{matrix}\right.\\
\forall x^{\theta}\in\Omega(q^2):\left(\rho^2\right)^{\rm A}(y|x^{\theta},p,q^2,f^c_p)&=\left\{\begin{matrix}
\theta/90, &  c\in [0,90]\wedge y=0 \text{ or }  c\in [90,180]\wedge y=1\\
 1-\theta/90, & c\in [0,90]\wedge y=1 \text{ or }  c\in [90,180]\wedge y=0\\
1,&   c\in [90, 180]\wedge y=0 \text{ or } c\in [270, 360]\wedge y=1\\
0,&   c\in [90, 180]\wedge y=1 \text{ or } c\in [270, 360]\wedge y=0\\
\end{matrix}\right.
\end{align*}
where $x^{\theta}:=(\cos\theta^{\circ},\sin\theta^{\circ})$.
Then, the overall target domain risk of the optimal learner for UDA classes $\pi^1$ and $\pi^2$ are:
\begin{align*}
R^{1*}_{\infty} &= 1-\mathbb{E}_{(p,q,f)\sim  \pi^1}\mathbb{E}_{x\sim q} [\max_{y} \left(\rho^1\right)^{\rm A}(y|x,p,q,f_p)]\\
&=1-\int_{0}^{360} \frac{1}{360}\int_{90}^{180}\frac{1}{90}\max_{y} \left(\rho^1\right)^{\rm A}(y|x^{\theta},p,q,f^c_p)d\theta d c\\
&=0\\
R^{2*}_{\infty} &= 1-\mathbb{E}_{(p,q,f)\sim  \pi^1}\mathbb{E}_{x\sim q} [\max_{y} \left(\rho^2\right)^{\rm A}(y|x,p,q,f_p)]\\
&=0.5-2\int_{0}^{90} \frac{1}{360}\int_{0}^{90}\frac{1}{90}\max_{y} \left(\rho^2\right)^{\rm A}(y|x^{\theta},p,q,f^c_p)d\theta d c\\
&=0.125
\end{align*}
\paragraph{Example 2.} The posterior distributions $\rho^1(\cdot|p,q,f^c)$ and $\rho^2(\cdot|q,p,f^c)$ are
\begin{align*}
    \rho^1(f^a|p,q,f^c_p)&=\left\{\begin{matrix}
1, &  a=c\\
0, &  \text{otherwise}\\
\end{matrix}\right.\\
\rho^2(f^a|q,p,f^c_q)&=\left\{\begin{matrix}
1/90, &  a\in [-45,45]\\
0, &  \text{otherwise}\\
\end{matrix}\right.
\end{align*}
The aggregated versions of  posterior for UDA classes $\pi^1$ and $\pi^2$ are:
\begin{align*}
\forall x\in\Omega(q):\left(\rho^1\right)^{\rm A}(y|x,p,q,f^c_p)&=\left\{\begin{matrix}
1, &  y = f^c(x) \\
 0, & y\neq f^c(x) \\
\end{matrix}\right.\\
\forall x^{\theta}\in\Omega(p):\left(\rho^2\right)^{\rm A}(1|x^{\theta};q,p,f^c_q)&=\left\{\begin{matrix}
(45+\theta)/90, &    \theta\in [0,45]\\
(225-\theta)/90, & \theta\in [135,225]\\
(\theta-315)/90,&  \theta\in [315,360]\\
\end{matrix}\right.
\end{align*}

where $x^{\theta}:=(\cos\theta^{\circ},\sin\theta^{\circ})$ and $\left(\rho^2\right)^{\rm A}(0|x^{\theta};q,p,f^c_p)=1-\left(\rho^2\right)^{\rm A}(1|x^{\theta};q,p,f^c_p)$. Then, the overall target domain risk of the optimal learner for UDA classes $\pi^1$ and $\pi^2$ are:
\begin{align*}
R^{1*}_{\infty} &=1-\int_{-45}^{45} \frac{1}{90}\int_{\Omega(q)}\frac{1}{180}\cdot 1 \;dx d c=0\\
R^{2*}_{\infty} &= 1-\int_{-45}^{45} \frac{1}{90}\left\{\int_{0}^{45}\frac{1}{180}M\left(\frac{45+\theta}{90}\right) d\theta +\int_{135}^{225}\frac{1}{180}M\left(\frac{225-\theta}{90} \right)d\theta +\int_{315}^{360}\frac{1}{180}M\left(\frac{\theta-315}{90} \right) d\theta \right\} dc\\
&=0.25
\end{align*}
where $M(\epsilon)=\max\{\epsilon, 1-\epsilon\}$ for any $\epsilon\in[0,1]$.
\paragraph{Example 3.} 
The posterior distributions $\rho^1(\cdot|p^1,q,f^c)$ and $\rho^2(\cdot|p^2,q,f^c)$ are
\begin{align*}
    \rho^1(f^a|p^1,q,f^c_{p^1})&=\left\{\begin{matrix}
1/2, &  a\in[-1,1]\\
0, &  \text{otherwise}\\
\end{matrix}\right.\\
\rho^2(f^a|p^2,q,f^c_{p^2})&=\left\{\begin{matrix}
1, &  c\in[-0.5,0.5]\wedge a\in [-0.5,0.5]\\
0, &  c\in[-0.5,0.5]\wedge a\notin [-0.5,0.5]\\
1, &  c\in[-1,-0.5]\cup [0.5,1]\wedge a=c\\
0, &  c\in[-1,-0.5]\cup [0.5,1]\wedge a\neq c\\
\end{matrix}\right.
\end{align*}
When $y=1$, the aggregated versions of  posterior for UDA classes $\pi^1$ and $\pi^2$ are:
\begin{align*}
\forall x\in\Omega(q):\left(\rho^1\right)^{\rm A}(1|x,p^1,q,f^c_{p^1})&=(1-x)/2\\
\forall x^{\theta}\in\Omega(q):\left(\rho^2\right)^{\rm A}(1|x,p^2,q,f^c_{p^2})&=\left\{\begin{matrix}
0, &   c\in[-0.5,0.5]\wedge a\in [-1,-0.5]\\
1, &   c\in[-0.5,0.5]\wedge a\in [0.5,1]\\
0.5-x, &c\in[-0.5,0.5]\wedge a\in [-0.5, 0.5]\\
0,&  c\in[-1,-0.5]\wedge a<c\\
1, &  c\in[-1,-0.5]\wedge a\geq c\\
0,&  c\in[0.5,1]\wedge a\geq c\\
1, &  c\in[0.5,1]\wedge a< c\\
\end{matrix}\right.
\end{align*}
The probability for $y=0$ is then obtained. Then, the overall target domain risk of the optimal learner for UDA classes $\pi^1$ and $\pi^2$ are:
\begin{align*}
R^{1*}_{\infty} &=1-\int_{-1}^{1} \frac{1}{2}\int_{-1}^{1}\frac{1}{2} M\left(\frac{1-x}{2}\right) dx d c=0.25\\
R^{2*}_{\infty} &= 1-\int_{-0.5}^{0.5} \frac{1}{2}\int_{-0.5}^{0.5}\frac{1}{2}M(0.5-x)dx  dc -\int_{-0.5}^{0.5}\frac{1}{2}\int_{[-1,-0.5]\cup[0.5,1]}\frac{1}{2}dxdc-2\int_{-1}^{-0.5}\frac{1}{2}\int_{-1}^1\frac{1}{2}dxdc\\
&=0.0625
\end{align*}
\paragraph{Example 4. } The posterior distributions $\rho(\cdot|p,q^c,f^c)$ and $\rho(\cdot|p,q^c,\bar{f}^c)$ are
\begin{align*}
    \rho(g|p,q^c,f^c)=\left\{\begin{matrix}
1, &  g=f^c\\
0, &  g\neq f^c\\
\end{matrix}\right.\\
\rho(g|p,q^c,\bar{f}^c)=\left\{\begin{matrix}
1, &  g=\bar{f}^c\\
0, &  g\neq \bar{f}^c\\
\end{matrix}\right.\\
\end{align*}
The aggregated versions of  posterior is:
\begin{align*}
\forall x\in\Omega(q^c):\rho^{\rm A}(y|x,p,q,f^c_p)&=\left\{\begin{matrix}
1, &  y = f^c(x) \\
 0, & y\neq f^c(x) \\
\end{matrix}\right.\\
\rho^{\rm A}(y|x;q,p,\bar{f}^c_q)&=\left\{\begin{matrix}
1, &  y = \bar{f}^c(x) \\
 0, & y\neq \bar{f}^c(x) \\
\end{matrix}\right.
\end{align*}
Then, the optimal overall target domain risk of this UDA class is:
\[
R_\infty^*=1-1=0.
\]

\begin{table}
    \centering 
    \caption{The values of compared measures in four examples for corresponding UDA instances.}
    \begin{tabular}{c|c|c|c}
    \hline
         Example&  Measures&  $\pi_1$&  $\pi_2$\\
    \hline
        \multirow{6}*{1}&  $f$-divergence & $\infty$ &  $\infty$\\
        ~ &  Wasserstein distance& $180^{\circ}$ &  $90^{\circ}$\\
        ~ &  $\mathcal{H}\Delta\mathcal{H}$ divergence & 0 & 1 \\
        ~ &  $\mathcal{Y}$-divergence & 0 &  1\\
        ~ &  Transfer Exponent &  1 &  $\infty$\\
       ~  &  PTLU & 0  & 0.5 \\
       \hline
        \multirow{6}*{2}&  $f$-divergence & $\infty$ &  $\infty$\\
        ~ &  Wasserstein distance& $90^{\circ}$ &  $90^{\circ}$\\
        ~ &  $\mathcal{H}\Delta\mathcal{H}$ divergence & 1 & 1 \\
        ~ &  $\mathcal{Y}$-divergence & 1/2 &  1/2\\
        ~ &  Transfer Exponent &  1 &  $\infty$\\
       ~  &  PTLU & 0  & 0.5 \\
       \hline
        \multirow{6}*{3}&  $f$-divergence & $\infty$ &  $\infty$\\
        ~ &  Wasserstein distance & lager &  smaller\\
        ~ &  $\mathcal{H}\Delta\mathcal{H}$ divergence & 1 & 1 \\
        ~ &  $\mathcal{Y}$-divergence & 1/2 &  1/2\\
        ~ &  Transfer Exponent &  1 &  $\infty$\\
       ~  &  PTLU & 0  & 0.5 \\
       \hline
        \multirow{5}*{4}&  $f$-divergence & \multicolumn{2}{c}{$\infty$}\\
        ~ &  $\mathcal{H}\Delta\mathcal{H}$ divergence & \multicolumn{2}{c}{1}\\
        ~ &  $\mathcal{Y}$-divergence & \multicolumn{2}{c}{1/2}\\
        ~ &  Transfer Exponent &  \multicolumn{2}{c}{$\infty$}\\
       ~  &  PTLU & \multicolumn{2}{c}{0}\\
        \hline
    \end{tabular}
    
    \label{tab:values of examples}
\end{table}

\section{Proof of Theorem \ref{thm:finite to infinite wrt xt and q}}
\label{proof:thm:finite to infinite wrt xt and q}
For any $x\in\Omega(q)$, we have:
\[
    0\leq \mathscr{H}(\rho^A(\cdot|x,s_{m,n}))\leq\log k
\]
By Hoeffding inequality, for any $t>0$, with probability at least $1-\exp\left(-\dfrac{2nt^2}{(\log k)^2}\right)$, in the sampling process of $\xt\sim(q)^n$, the following holds:
\begin{equation}
    \Big\vert
    U(s_{m,n})-\widetilde{U}(s_{m,n})
    \Big\vert
    =\Big\vert
    \mathbb{E}_{x\sim q}\mathscr{H}\left(\rho^{\rm A}(\cdot|x,s_{m,n})\right)
    -
    \dfrac{1}{n}\sum_{x\in\xt}\mathscr{H}\left(\rho^{\rm A}(\cdot|x,s_{m,n})\right)
    \Big\vert\leq
    t
    \label{eq:hoeffding inequality}
\end{equation}
Combining Eq. \ref{eq:hoeffding inequality} and Theorem \ref{thm:LB_mn}
gives us Theorem \ref{thm:finite to infinite wrt xt and q}.

\section{Proof of Theorem \ref{thm:finite to infinite}}
\label{proof:thm:finite to infinite}
Before proving Theorem \ref{thm:finite to infinite}, we first introduce some concepts/notions that will be used later.
\\
Let $\mathcal{A}$ be a finite alphabet. For any $\mu_1,\mu_2\in\Delta(\mathcal{A})$, the Hellinger distance $h(\mu_1,\mu_2)$ is defined as:
\[
    h(\mu_1,\mu_2):=\dfrac{1}{\sqrt{2}}\sqrt{\mathop{\sum_{a\in\mathcal{A}}}\Big(\small\sqrt{\mu_1(a)}-\small\sqrt{\mu_2(a)}\Big)^2},
\]
and the total variance distance (TVD) $d(\mu_1,\mu_2)$ is defined as:
\[
d(\mu_1,\mu_2):=\dfrac{1}{2}\sum_{a\in\mathcal{A}}\Big\vert{\mu_1(a)-\mu_2(a)}\Big\vert.
\]
Again for any $\mu_1,\mu_2\in\Delta(\mathcal{A})$, we define 
\[
\bigg\Vert\dfrac{\mu_1}{\mu_2}\bigg\Vert_\infty := \underset{x\in\mathcal{X}}{\max}\bigg\vert{\dfrac{\mu_1(x)}{\mu_2(x)}}\bigg\vert,
\]
and let $\mu_1\otimes\mu_2:\mathcal{A}\times\mathcal{A}\rightarrow[0,1]$ denote the distribution induced by the product of $\mu_1$ and $\mu_2$, that is:
\[
\forall a_1,a_2\in\mathcal{A},\;\;\;\mu_1\otimes\mu_2(a_1,a_2)=\mu_1(a_1)\cdot\mu_2(a_2).
\]

\begin{lem}
     Define $r={m\wedge n}$. Suppose for a sequence $(\epsilon_i)$ of positive numbers satisfying
     \begin{align*}
         &\underset{i\rightarrow\infty}{\lim}\epsilon_i=0\\
         &\underset{i\rightarrow\infty}{\lim}i\epsilon_i^2=\infty\\
         &\underset{i\rightarrow\infty}{\lim}\sum_{j=1}^i\exp(-j\epsilon_j^2B)<\infty\quad\quad\text{for any $B>0$},
     \end{align*}
     a constant $C>0$, and a sequence $(\mathcal{E}_i)$ of subsets $\mathcal{E}_i\subseteq\mathcal{E}$ we have that:
     \begin{align*}
        &\log \mathcal{M}(\epsilon_r,\mathcal{E}_i,d_\text{TV})\leq i\epsilon_i^2\\
        &\pi_{P,Q}\Big(\mathcal{E}\backslash\mathcal{E}_i\Big)\leq \exp\big(-i
        \epsilon  _i^2(C+4)\big)\\
        &\pi_{P,Q}\Big({(p',q'):h^2\big(p'\otimes q',p\otimes q\big)\cdot\bigg\Vert{\dfrac{p'\otimes q'}{p\otimes q}}\bigg\Vert_\infty}\leq \epsilon_i^2\Big)\geq \exp\big(-i\epsilon_i^2 C\big)
    \end{align*}
   Define the posterior of $(P,Q)$ upon observing $(\xs,\xt)$ by $\pi_{P,Q|X_{\rm s},X_{\rm t}}(\cdot,\cdot|\xs,\xt)$.
   
   Then the following inequality holds:
   \begin{align*}
       \underset{\substack{\text{\bf x}_\text{s} \sim (p)^m \\ \text{\bf x}_\text{t} \sim (q)^n}}{\Pr}\bigg\{\pi_{P,Q|X_\text{s},X_\text{t}}\Big(p',q':d_\text{TV}(p'\otimes q',p\otimes q)>M\epsilon_r\Big|\xs,\xt\Big)\leq\exp(-B_2r\epsilon_r^2)\bigg\}\geq 1-\exp(-B_1r\epsilon_r^2)
   \end{align*}
   for $M\geq\sqrt{\dfrac{C+4}{C_1}}\quad\text{for some }C_1>0$.
    \label{lem:convergence rates of joint posterior}
\end{lem}
\begin{rem}
    Lemma \ref{lem:convergence rates of joint posterior}  is derived from Theorem 2.1 of \cite{ghosal2000convergence}, which describes the convergence behavior of the posterior. Specifically, it implies that as long as the prior $\pi_{P,Q}$  puts sufficient probability mass on $(p,q)$ and its close neighbors (in the sense of their TVD), then the probability that the posterior will also sufficiently put the majority of its probability mass near $(p,q)$ converges to $1$ at exponential rate. 
\end{rem}
Let the assumptions in Lemma \ref{lem:convergence rates of joint posterior}
 hold. For any $f'\in\mathcal{F}$, we also assume that $\pi_{F|P,Q}(f'|p,q)$, as a function of $(p,q)$, is $L$-Lipschitz with respect to $d_\text{TV}$, namely:
\begin{align*}
 \Big\vert\mathcal{G}_f(p_1,q_1)-\mathcal{G}_f(p_2,q_2)\Big\vert\leq L\cdot d\Big({(p_1,q_1),(p_2,q_2)}\Big)
\end{align*}
for any $(p_1,q_1),(p_2,q_2)\in\Delta(\mathcal{X})\times\Delta(\mathcal{X})$. Denote the event 
\begin{align*}
  \pi_{P,Q|X_{\rm s},X_{\rm t}}\Big(p',q':d_\text{TV}(p'\otimes q',p\otimes q)>
  M\epsilon_r\Big|\xs,\xt\Big)\leq\exp(-B_2r\epsilon_r^2)
\end{align*}
 by $E_1$. Then if $E_1$ holds true, we have that:
 \begin{align*}
    &\big\vert
     \pi_{F|X_\text{s},X_\text{t}}(f'|\xs,\xt)-\pi_{F|P,Q}(f'|p,q)
     \big\vert\\
     :=&\Bigg\vert
    \int_{(p',q')} \pi_{F|P,Q}(f'|p',q')\pi_{P,Q|X_\text{s},X_\text{t}}(p',q'|\xs,\xt)d(p',q')-\pi_{F|P,Q}(f'|p,q)
     \Bigg\vert\\
     \leq&\int_{(p',q')}\bigg\vert
     \pi_{F|P,Q}(f'|p',q')-\pi_{F|P,Q}(f'|p,q)
     \bigg\vert \pi_{P,Q|X_\text{s},X_\text{t}}(p',q'|\xs,\xt)d(p',q')\\
     =&\int_{(p',q'):d_\text{TV}(p'\otimes q',p\otimes q)\leq M\epsilon_r^2}\bigg\vert
     \pi_{F|P,Q}(f'|p',q')-\pi_{F|P,Q}(f'|p,q)
     \bigg\vert \pi_{P,Q|X_\text{s},X_\text{t}}(p',q'|\xs,\xt)d(p',q')\\
     &+\int_{(p',q'):d_\text{TV}(p'\otimes q',p\otimes q)> M\epsilon_r^2}\bigg\vert
     \pi_{F|P,Q}(f'|p',q')-\pi_{F|P,Q}(f'|p,q)
     \bigg\vert \pi_{P,Q|X_\text{s},X_\text{t}}(p',q'|\xs,\xt)d(p',q')\\
     \leq&
      LM\epsilon_r\int_{(p',q'):d_\text{TV}(p'\otimes q',p\otimes q)\leq M\epsilon_r^2}
     \pi_{P,Q|X_\text{s},X_\text{t}}(p',q'|\xs,\xt)d(p',q')\\
     &+\int_{(p',q'):d_\text{TV}(p'\otimes q',p\otimes q)> M\epsilon_r^2}
     \pi_{P,Q|X_\text{s},X_\text{t}}(p',q'|\xs,\xt)d(p',q')\\
     \leq&
     LM\epsilon_r+\exp(-r\epsilon_r^2B_2)\\
     \leq&
     B_3\epsilon_r
 \end{align*}
 
 for a constant $B_3=\mathcal{O}(L)$ and each $f'\in\mathcal{F}$. Then we have that:
 \begin{align*}
     \Pr\bigg\{
     \Big\vert
     \pi_{F|X_\text{s},X_\text{t}}(f'|\xs,\xt)-\pi_{F|P,Q}(f'|p,q)
     \Big\vert
     \leq B_3\epsilon_r
     \bigg\}
     \geq
     \Pr\{E_1\}
     \geq
     1-\exp(-r\epsilon_r^2B_2)
 \end{align*}
for each $f'\in\mathcal{F}$.\\
\\
We assume that the set $\mathcal{F}[f_p]$ is finite, i.e. $K<\infty$. In the following context, we condition our statement on the assumption that $\mathcal{F}[\xs,\text{\bf y}_\text{s}]=\mathcal{F}[f_p]$, then we have that both $\rho_\infty(\cdot|p,q,f_p)$ and $\rho(\cdot|s_{m,n})$ are finite-dimensional discrete distribution on set $\mathcal{F}[f_p]$. Define $S':=\underset{f'\in\mathcal{F}[f_p]}{\sum}\pi_{F|X_\text{s},X_\text{t}}(f'|\xs,\xt)$. Then if the inequality
\begin{align*}
    \Big\vert
     \pi_{F|X_\text{s},X_\text{t}}(f'|\xs,\xt)-\pi_{F|P,Q}(f'|p,q)
     \Big\vert
     \leq B_3\epsilon_r
\end{align*}
holds true for all $f'\in\mathcal{F}[f_p]$, we have that, for each $f'\in\mathcal{F}[f_p]$:
\begin{align*}
    &\Big\vert
    \rho_\infty(f'|p,q,f_p)-\rho(f'|s_{m,n})
    \Big\vert\\
   :=&\Bigg\vert
   \dfrac{\pi_{F|P,Q}(f'|p,q) }{S}-\dfrac{\pi_{F|X_\text{s},X_\text{t}}(f'|\xs,\xt)}{S'}
   \Bigg\vert\\
   =&
   \Bigg\vert
    \dfrac{\pi_{F|P,Q}(f'|p,q)\cdot S'-\pi_{F|X_\text{s},X_\text{t}}(f'|\xs,\xt)\cdot S}{SS'}
   \Bigg\vert\\
   =&
   \Bigg\vert
    \dfrac{\pi_{F|P,Q}(f'|p,q)(S'-S)+S\Big(\pi_{F|P,Q}(f'|p,q)-\pi_{F|X_\text{s},X_\text{t}}(f'|\xs,\xt)\Big)}{SS'}
   \Bigg\vert\\
   \leq&
   \dfrac{\pi_{F|P,Q}(f'|p,q)|S-S'|+S\Big\vert\pi_{F|P,Q}(f'|p,q)-\pi_{F|X_\text{s},X_\text{t}}(f'|\xs,\xt)\Big\vert}{SS'}\\
   \leq&
   \dfrac{\pi_{F|P,Q}(f'|p,q)\cdot KB_3\epsilon_r+ SB_3\epsilon_r}{S\cdot|S-KB_3\epsilon_r|}\\
   \leq&
   \dfrac{(K+S)B_3\epsilon_r}{S\cdot|S-KB_3\epsilon^2|}\\
   \leq&
   B_4\epsilon_r
\end{align*}

for a constant $B_4=\mathcal{O}\bigg(\dfrac{KL}{S^2}\bigg)$. Therefore, for each $f'\in\mathcal{F}[f_p]$, we have that:
\begin{equation}
\begin{aligned}
    &\Pr\bigg\{
    \Big\vert
    \rho_\infty(f'|p,q,f_p)-\rho(f'|s_{m,n})
    \Big\vert
    \leq B_4\epsilon_r\bigg\}\\
    \geq&
    \Pr\bigg\{
     \Big\vert
     \pi_{F|X_\text{s},X_\text{t}}(f''|\xs,\xt)-\pi_{F|P,Q}(f''|p,q)
     \Big\vert
     \leq B_3\epsilon_r\quad\forall f''\in\mathcal{F}[f_p]
    \bigg\}\\
    \geq&
    \underset{f''\in\mathcal{F}[f_p]}{\sum}\Pr\bigg\{
     \Big\vert
     \pi_{F|X_\text{s},X_\text{t}}(f''|\xs,\xt)-\pi_{F|P,Q}(f''|p,q)
     \Big\vert
     \leq B_3\epsilon_r
    \bigg\}-\big|\mathcal{F}[f_p]\big|+1\\
    \geq&
    1-K\exp(-r\epsilon_r^2B_2)
\label{eq:rho finite vs infinite}
\end{aligned}
\end{equation}
For each given $x\in\Omega(q)$, define subsets $I_x^1,\cdots,I_x^k$ such that:
\begin{align*}
    &I_x^i\neq\emptyset\ \forall i=1,\cdots,k\\
    &I_x^i\cap I_x^j=\emptyset\ \ \ \ \forall i\neq j\\
    &I_x^1\cup\cdots\cup I_x^k=\mathcal{F}[f_]\\
    &f(x)=i\ \ \ \ \forall f\in I_x^i\ \text{and}\ i=1,\cdots,k
\end{align*}

 For each $y\in\mathcal{Y}$, assume the following inequality holds true for all $f'\in I_x^y$:
 \begin{align*}
      \Big\vert
    \rho_\infty(f'|p,q,f_p)-\rho(f'|s_{m,n})
    \Big\vert
    \leq B_4\epsilon_r
 \end{align*}
 Then we have that:
 \begin{align*}
     &\bigg\vert
     \rho^\text{A}(y|x,p,q,f_p)-\rho^\text{A}(y|x,s_{m,n})
     \bigg\vert\\
     =&\Bigg\vert
     \underset{f'\in I_x^y}{\sum}\Big(
        \rho_\infty(f'|p,q,f_p)-\rho(f'|s_{m,n})
        \Big)
     \Bigg\vert\\
     \leq&
     \underset{f'\in I_x^y}{\sum}\Big\vert
     \rho_\infty(f'|p,q,f_p)-\rho(f'|s_{m,n})
     \Big\vert\\
     \leq&
     \big\vert I_x^y \big\vert B_4\epsilon_r
 \end{align*}
 and we will also have that:
 \begin{align*}
     &\Bigg\vert
     \rho^\text{A}(y|x,p,q,f_p)\log \rho^\text{A}(y|x,p,q,f_p)-\rho^\text{A}(y|x,s_{m,n})\log\rho^\text{A}(y|x,s_{m,n})
     \Bigg\vert\\
     \leq&
     \Bigg\vert
     \beta\log\beta-(\beta+\big\vert I_x^y \big\vert B_4\epsilon_r)\log(\beta+\big\vert I_x^y \big\vert B_4\epsilon_r)
     \Bigg\vert\\
     \leq&
     B_5 \big\vert I_x^y \big\vert B_4\epsilon_r
 \end{align*}
 for a constant $B_5=\mathcal{O}\Big(\log(1/\beta)-1\Big)$. We may write $B_5 \big\vert I_x^y \big\vert B_4\epsilon_r$ as $\big\vert I_x^y \big\vert B_6\epsilon_r$ for a constant $B_6=\mathcal{O}\bigg(\dfrac{KL(\log(1/\beta))}{S^2}\bigg)$. Then, for each $y\in\mathcal{Y}$, we have that:
 \begin{align*}
     &\Pr\Bigg\{
     \Bigg\vert
     \rho^\text{A}(y|x,p,q,f_p)\log \rho^\text{A}(y|x,p,q,f_p)-\rho^\text{A}(y|x,s_{m,n})\log\rho^\text{A}(y|x,s_{m,n})
     \Bigg\vert\leq\big\vert I_x^y \big\vert B_6\epsilon_r
     \Bigg\}\\
     \geq&
     \Pr\Bigg\{
     \Big\vert
    \rho_\infty(f'|p,q,f_p)-\rho(f'|s_{m,n})
    \Big\vert
    \leq B_4\epsilon_r\quad\forall f'\in I_y^x
     \Bigg\}\\
     \geq&\underset{f'\in I_y^x}{\sum}\Pr\Bigg\{
     \Big\vert
    \rho_\infty(f'|p,q,f_p)-\rho(f'|s_{m,n})
    \Big\vert
    \leq B_4\epsilon_r
     \Bigg\}+\big\vert I_x^y \big\vert-1\\
     \geq&
     1-\big\vert I_x^y \big\vert K\exp(-r\epsilon_r^2B_2).
 \end{align*}
Assume for each $x\in\Omega(q)$, the following inequality holds for all $y\in\mathcal{Y}$:
\begin{align*}
    \Bigg\vert
     \rho^\text{A}(y|x,p,q,f_p)\log \rho^\text{A}(y|x,p,q,f_p)-\rho^\text{A}(y|x,s_{m,n})\log\rho^\text{A}(y|x,s_{m,n})
     \Bigg\vert\leq\big\vert I_x^y \big\vert B_6\epsilon_r
\end{align*}
then we have that:
\begin{align*}
    &\bigg\vert
    \mathscr{H}\Big(\rho^\text{A}(\cdot|x,p,q,f_p)\Big)-\mathscr{H}\Big(\rho^\text{A}(\cdot|x,s_{m,n})\Big)
    \bigg\vert\\
    =&
    \Bigg\vert
    \underset{y\in\mathcal{Y}}{\sum}-\rho^\text{A}(y|x,p,q,f_p)\log \rho^\text{A}(y|x,p,q,f_p)-\underset{y\in\mathcal{Y}}{\sum}-\rho^\text{A}(y|x,s_{m,n})\log\rho^\text{A}(y|x,s_{m,n})
    \Bigg\vert\\
    \leq&
    \underset{y\in\mathcal{Y}}{\sum}\big\vert I_x^y \big\vert B_6\epsilon_r\\
    =&KB_6\epsilon_r\\
    =&B_7\epsilon_r
\end{align*}
for a constant $B_7=\mathcal{O}\bigg(\dfrac{K^2L\log(1/\beta)}{S^2}\bigg)$. This will give us that, for each $x\in\Omega(q)$:
\begin{align*}
    &\Pr\Bigg\{
    \bigg\vert
    \mathscr{H}\Big(\rho^\text{A}(\cdot|x,p,q,f_p)\Big)-\mathscr{H}\Big(\rho^\text{A}(\cdot|x,s_{m,n})\Big)
    \bigg\vert\leq B_7\epsilon_r
    \Bigg\}\\
    \geq&\Pr\Bigg\{
     \Bigg\vert
     \rho^\text{A}(y|x,p,q,f_p)\log \rho^\text{A}(y|x,p,q,f_p)-\rho^\text{A}(y|x,s_{m,n})\log\rho^\text{A}(y|x,s_{m,n})
     \Bigg\vert\leq\big\vert I_x^y \big\vert B_6\epsilon_r\quad\forall y\in\mathcal{Y}
    \Bigg\}\\
    \geq&
    \underset{y\in\mathcal{Y}}{\sum}\Pr\Bigg\{
    \Bigg\vert
     \rho^\text{A}(y|x,p,q,f_p)\log \rho^\text{A}(y|x,p,q,f_p)-\rho^\text{A}(y|x,s_{m,n})\log\rho^\text{A}(y|x,s_{m,n})
     \Bigg\vert\leq\big\vert I_x^y \big\vert B_6\epsilon_r
    \Bigg\}-k+1\\
    \geq&
    1-K^2\exp(-r\epsilon_r^2B_2)
\end{align*}
From now on, let's also assume that all points in $\Omega(q)$ have been sampled in $\xs$. Then if the following inequality holds for all $x\in\Omega(q)$:
\begin{align*}
    \bigg\vert
    \mathscr{H}\Big(\rho^\text{A}(\cdot|x,p,q,f_p)\Big)-\mathscr{H}\Big(\rho^\text{A}(\cdot|x,s_{m,n})\Big)
    \bigg\vert\leq B_7\epsilon_r
\end{align*}
we will have that:
\begin{align*}
    &\Bigg\vert
    U(p,q,f_p)-\widetilde{U}(s_{m,n})
    \Bigg\vert\\
    =&
    \Bigg\vert
    \mathbb{E}_{x\sim q}\mathscr{H}\Big(\rho^\text{A}(\cdot|x,p,q,f_p)\Big)-\mathbb{E}_{x\sim\xt}\mathscr{H}\Big(\rho^\text{A}(\cdot|x,s_{m,n})\Big)
    \Bigg\vert\\
    =&\Bigg\vert
    \mathbb{E}_{x\sim q}\mathscr{H}\Big(\rho^\text{A}(\cdot|x,p,q,f_p)\Big)-\mathbb{E}_{x\sim q}\mathscr{H}\Big(\rho^\text{A}(\cdot|x,s_{m,n})\Big)
    \Bigg\vert\\
    \leq&
    \mathbb{E}_{x\sim q}\Bigg\vert
    \mathscr{H}\Big(\rho^\text{A}(\cdot|x,p,q,f_p)\Big)-\mathscr{H}\Big(\rho^\text{A}(\cdot|x,s_{m,n})\Big)
    \Bigg\vert\\
    \leq&
    \mathbb{E}_{x\sim q}[B_7\epsilon_r]\\
    =&B_7\epsilon_r
\end{align*}
Then overall, under the conditions that 1. $\mathcal{F}[f_p]=\mathcal{F}[\xs,\text{\bf y}_\text{s}]$, and 2. $\underset{x\in\Omega(q)}{\bigcap}\{x\in\xt\}$, we have:
\begin{align*}
    &Pr\Bigg\{
    \Bigg\vert
    U(p,q,f_p)-\widetilde{U}(s_{m,n})
    \Bigg\vert\leq B_7\epsilon_r
    \Bigg\}\\
    \geq&
    \Pr\Bigg\{
     \bigg\vert
    \mathscr{H}\Big(\rho^\text{A}(\cdot|x,p,q,f_p)\Big)-\mathscr{H}\Big(\rho^\text{A}(\cdot|x,s_{m,n})\Big)
    \bigg\vert\leq B_7\epsilon_r\quad x\in\Omega(q)
    \Bigg\}\\
    \geq&
    \underset{x\in\Omega(a)}{\sum}\Pr\Bigg\{
     \bigg\vert
    \mathscr{H}\Big(\rho^\text{A}(\cdot|x,p,q,f_p)\Big)-\mathscr{H}\Big(\rho^\text{A}(\cdot|x,s_{m,n})\Big)
    \bigg\vert\leq B_7\epsilon_r
    \Bigg\}-\Big\vert \Omega(q) \Big\vert+1\\
    \geq&
    1-N_qK^2\exp(-r\epsilon_r^2B_2)
\end{align*}
Next, we will quantify the probabilities of the two conditions. We have assumed that $m>N_p$. Thus, we have that:
\begin{align*}
    &\Pr\Bigg\{
    \underset{x\in\Omega(p)}{\bigcup}\{x\notin\xs\}
    \Bigg\}\\
    \leq&\sum_{i=1}^{N_p}\Pr\Big\{
        x_i\notin\xs
    \Big\}\\
    \leq&
    N_p(1-\alpha_p)^m\\
    \leq&
    N_p\exp(-m\alpha_p)
\end{align*}
Then:
\begin{align*}
    \Pr\Big\{
    \mathcal{F}[f_p]=\mathcal{F}[\xs,\text{\bf y}_\text{s}]
    \Big\}
    \geq
    \Pr\Bigg\{
    \underset{x\in\Omega(p)}{\bigcap}\{x\in\xs\}
    \Bigg\}\geq 1-N_p\exp(-m\alpha_p)
\end{align*}
Similarly, we can show that:
\begin{align*}
     \Pr\Bigg\{
    \underset{x\in\Omega(q)}{\bigcap}\{x\in\xt\}
    \Bigg\}\geq 1-N_q\exp(-n\alpha_q)
\end{align*}
Therefore, without specifically assuming the $2$ conditions to hold, in general we have that:
\begin{equation}
\begin{aligned}
    Pr\Bigg\{
    \Bigg\vert
    U(p,q,f_p)-\widetilde{U}(s_{m,n})
    \Bigg\vert\leq B_7\epsilon_r
    \Bigg\}&\geq \Big(1-N_qK^2\exp(-r\epsilon_r^2B_2)\Big)\Big(1-\exp(-m\alpha_p)\Big)\Big(1-\exp(-n\alpha_q)\Big)\\
    &\geq 1-N_qK^2\exp\left(-r\epsilon_r^2B_2\right)-N_p\exp(-m\alpha_p)-N_q\exp(-n\alpha_q)
    \label{eq:U vs U tilde}
\end{aligned}
\end{equation}
Combining Eq. \ref{eq:U vs U tilde} and Corollary \ref{coro:LB_infty} will give us Theorem \ref{thm:finite to infinite}.
$\hfill\square$

\end{document}